\documentclass[energies,article,accept,moreauthors,pdftex]{Definitions/mdpi}

\firstpage{1}
\pubvolume{14}
\issuenum{12}
\articlenumber{3623}
\pubyear{2021}
\copyrightyear{2021}
\externaleditor{Academic Editors: Djaffar Ould-Abdeslam, Igor Simone Stievano and Riccardo Trinchero} % Please add Academic Editor name
\datereceived{11 May 2021}
\dateaccepted{16 June 2021}
\datepublished{18 June 2021}
\hreflink{https://doi.org/10.3390/en14123623} % If needed use \linebreak
\Title{Deep Learning for High-Impedance Fault Detection: Convolutional Autoencoders}

\TitleCitation{Deep Learning for High-Impedance Fault Detection: Convolutional Autoencoders}

\Author{Khushwant Rai 
\href{https://orcid.org/0000-0001-8068-7794}{\orcidicon}, Farnam Hojatpanah \href{https://orcid.org/0000-0001-5657-4669}{\orcidicon}, Firouz Badrkhani Ajaei \href{https://orcid.org/0000-0002-2635-4128}{\orcidicon} and Katarina Grolinger *\href{https://orcid.org/0000-0003-0062-8212}{\orcidicon}}

\AuthorNames{Khushwant Rai, Farnam Hojatpanah, Firouz Badrkhani Ajaei and Katarina Grolinger}

\AuthorCitation{Rai, K.; Hojatpanah, F.; Badrkhani Ajaei, F.; Grolinger, K.}
\address[1]{Department of Electrical and Computer Engneering, The University of Western Ontario, London,~ON~N6A~5B9,~Canada; kkhushwa@uwo.ca (K.R.); fhojatpa@uwo.ca (F.H.);  fajaei@uwo.ca (F.B.A.)}
\corres{\hangafter=1 \hangindent=1.05em \hspace{-0.82em}Correspondence: kgroling@uwo.ca; Tel.: +1-519-661-2111 (ext. 81407)}

\abstract{High-impedance faults (HIF) are difficult to detect because of their low current amplitude and highly diverse characteristics. In recent years, machine learning (ML) has been gaining popularity in HIF detection because ML techniques learn patterns from data and successfully detect HIFs. However, as these methods are based on supervised learning, they fail to reliably detect any scenario, fault or non-fault, not present in the training data. Consequently, this paper takes advantage of unsupervised learning and proposes a convolutional autoencoder framework for HIF detection (CAE-HIFD). Contrary to the conventional autoencoders that learn from normal behavior, the convolutional autoencoder (CAE) in CAE-HIFD learns only from the HIF signals eliminating the need for presence of diverse non-HIF scenarios in the CAE training. CAE distinguishes HIFs from non-HIF operating conditions by employing cross-correlation. To discriminate HIFs from transient disturbances such as capacitor or load switching, CAE-HIFD uses kurtosis, a statistical measure of the probability distribution shape. The performance evaluation studies conducted using the IEEE 13-node test feeder indicate that the CAE-HIFD reliably detects HIFs, outperforms the state-of-the-art HIF detection techniques, and is robust against noise.}


\begin{document}

\section{Introduction} \label{sec: intro}
High-impedance faults (HIF) typically occur when a live conductor contacts a highly resistive surface~\cite{pow1, pow2, pow3, pow5}. They commonly have characteristics such as asymmetry of the current waveform, randomness, and~non-linearity of the voltage-current relationship. These~characteristics are diverse and are affected by multiple factors including surface types and humidity conditions~\cite{pow5}. As~a result, the~HIF current magnitude typically ranges from 0 to 75 A~\cite{pow1}. Such low current magnitudes compared to normal load current levels together with a high diversity of characteristics and patterns make HIFs difficult to detect. Specifically, the~conventional overcurrent relays fail to discriminate most HIFs from load unbalance~\cite{pow1,pow2,pow7}. Nevertheless, the~arcing ignition caused by an HIF is a safety hazard~\cite{pow5, pow4}. Moreover, undetected HIFs have been reported to cause instability of renewable energy systems~\cite{wind}. As~a result, reliably detecting and clearing HIFs in a timely manner are crucial to ensure the safety of personnel and maintain the power system integrity~\cite{pow5, pow4, pow7}.

Various HIF detection techniques have been proposed. The~harmonic-based detection is one of the most common techniques~\cite{pow1, pow18, pow19, pow20, pow21, pow22}. It operates well when the measured signals contain a large amount of harmonics, but~it loses sensitivity when the HIF is far from the relay~\cite{pow1}. An~HIF detection scheme based on a power line communication system is proposed by Milioudis~et~al. \cite{pow28}. This method is successful in detecting and locating HIFs, but~requires costly communication systems and is not suitable for large and complex networks~\cite{pow27}.

The discrete wavelet transformation-based HIF detection technique~\cite{pow11} operates by examining the measured signals in both the time and frequency domains. Although this method reduces the HIF detection delay, its performance is highly dependent on the choice of the mother wavelet and is adversely affected by the presence of unbalanced loads and noise~\cite{pow26}.

HIF detection by Kalman Filtering~\cite{pow10} is based on tracking randomness in the current waveform. Because~the small HIF current does not cause significant variations in the current waveforms seen by the relay at the substation~\cite{pow29}, this approach exhibits low~sensitivity.

The approach based on mathematical morphology~\cite{pow2} detects close-in HIFs but is not sufficiently reliable in presence of non-linear loads and under remote HIFs~\cite{pow26}. Empirical mode decomposition (EMD) \cite{pow30} and variational mode decomposition (VMD) \cite{pow4, pow7} have been recently proposed for HIF detection. These techniques operate by extracting HIF-related features from the signals. However, the~EMD-based method suffers from modal mixing and is sensitive to noise~\cite{pow31}. On~the other hand, the~VMD-based technique is more robust against noise, nevertheless, it requires a large number of decomposition modes, as well as a high sampling frequency to perform well, which results in a large computational burden. Moreover, the~VMD technique utilized in~\cite{pow7} for HIFs detection cannot reliably detect HIFs when a large non-linear load is in~operation.

In recent years, machine learning (ML) approaches have been gaining popularity. A~support vector machine (SVM)-based model is used with features extracted by VMD for HIF detection in distribution lines incorporating distributed generators~\cite{pow4}. This method prevents false detection of switching events as HIFs; however, it is sensitive to noise for signal to noise ratios (SNR) below 30 dB. Moreover, the~transformations associated with the VMD require high computational capability~\cite{pow5}. An~artificial neural network (ANN) is combined with discrete wavelet transforms (DWT) for HIF detection in medium-voltage networks. The~DWTs are responsible for extracting the relevant features from the current waveforms, whereas the ANN acts as a classifier using those extracted features~\cite{ANN}. Probabilistic neural networks and various fuzzy inference systems have also been combined with wavelet transform (WT) for HIF detection~\cite{Fuzzy, pow13}. In~another study~\cite{CNN_hifd}, a~modified Gabor WT is applied on the input signal to extract features in the form of two dimensional (2-D) scalograms. The~classifiers such as adaptive neuro-fuzzy inference system and SVM are utilized along with DWT-based feature extraction to detect HIFs and to discriminate them from other transients in medium-voltage distribution systems {\cite{SVMFuzzy}}. The~neural networks are sensitive to frequency changes, whereas the WT faces challenges in terms of selecting the mother wavelet. The~decision tree algorithm with the fast Fourier transform is also an alternative approach, but~it is sensitive to noise~\cite{DecisionT}.

Recurrent neural network (RNN)-based long short term memory (LSTM) approach is used with features obtained by DWT analysis to detect the HIFs in the solar photovoltaic integrated power system {\cite{LSTM}}.
However, this method detects HIFs with a success rate of only  92.42\% and, moreover, this study uses  DWT-based features which are sensitive to noise {\cite{pow26}}. In~another study, the~HIFs were detected and classified with 2-D convolutional neural network (CNN) which employs supervised learning to learn from the extracted features~\cite{CNN_hifd}. This method was accurate in the presented experiments; however, as~it is a supervised learning-based method, it may fail to reliably detect HIF and non-HIF scenarios that are not present in the training~data.

All reviewed ML studies apply a supervised approaches to learn the mapping from the input to the output based on a limited set of HIF and non-HIF scenarios present in the training set. Here, the~non-HIF scenario refers to  normal steady-state operation and any transient response to non-HIF disturbance, such as capacitor switching and load variations. A~supervised learning system may fail to reliably identify any fault scenario or any non-fault disturbance which is not present in the training set. However, there is a wide range of non-fault operating conditions and it is difficult to include them all in the training~set. Therefore, a~different way of training ML models is required. Moreover, the~existing technical literature on ML-based HIF detection relies on features extracted by resource intensive signal processing and data transformation techniques~\cite{LHeureux2017}.

Consequently, this paper proposes the convolutional autoencoder (CAE) framework for HIF detection (CAE-HIFD), which utilizes an unsupervised approach that learns solely from the fault data, thus avoiding the need to take into account all possible non-fault scenarios in the learning stage. The~ability of the CAE to model the relationship between the data points constituting a signal enables the CAE-HIFD to learn complex HIF patterns. The~CAE discriminates steady-state conditions from HIFs by identifying deviations from the learned HIF patterns using cross-correlation (CC). The~security against false detection of non-HIF disturbances (e.g., capacitor and load switching) as HIFs is achieved through kurtosis~analysis.

The performance of the proposed protection strategy is evaluated through comprehensive studies conducted on the IEEE 13-node test feeder taking into account various HIF conditions involving seven different fault surfaces, as well as diverse non-HIF scenarios. The~results indicate that the proposed CAE-HIFD (i) reliably detects HIFs regardless of the type of the surface involved and the fault distance, (ii) accurately discriminates between HIFs, steady-state operating conditions, and non-HIF disturbances, (iii) achieves higher accuracy than the state-of-the-art HIF detection techniques, and~(iv) is robust against~noise.

The paper is organized as follows: Section~\ref{sec:AE} provides an overview of autoencoders, Section~\ref{sec:methodology} presents the proposed CAE-HIFD, Section~\ref{sec:results} evaluates the performance of the CAE-HIFD, and~finally Section~\ref{sec:conclusion} concludes the~paper.

\section{Unsupervised Learning with~Autoencoders} \label{sec:AE}
A supervised machine learning model learns the mapping function from the input to output (label) based on example input--output pairs provided in the training dataset. In~contrast, an~unsupervised learning model discovers patterns and learns from the input data on its own, without~the need for labeled responses, which makes this approach a go-to solution when labels are not available. Autoencoders are trained through unsupervised learning where the model learns data encoding by reconstructing the input~\cite{deeplearnbook}. They are commonly used for dimensionality reduction~\cite{deeplearnbook, ghosh2020}, but~the non-linear feature learning capability has made them also successful in denoising and anomaly detection~\cite{autoenc_anom, araya2017ensemble, AEdenoise}.

\begin{figure}[H]
%\centering
\includegraphics[width=8.5 cm]{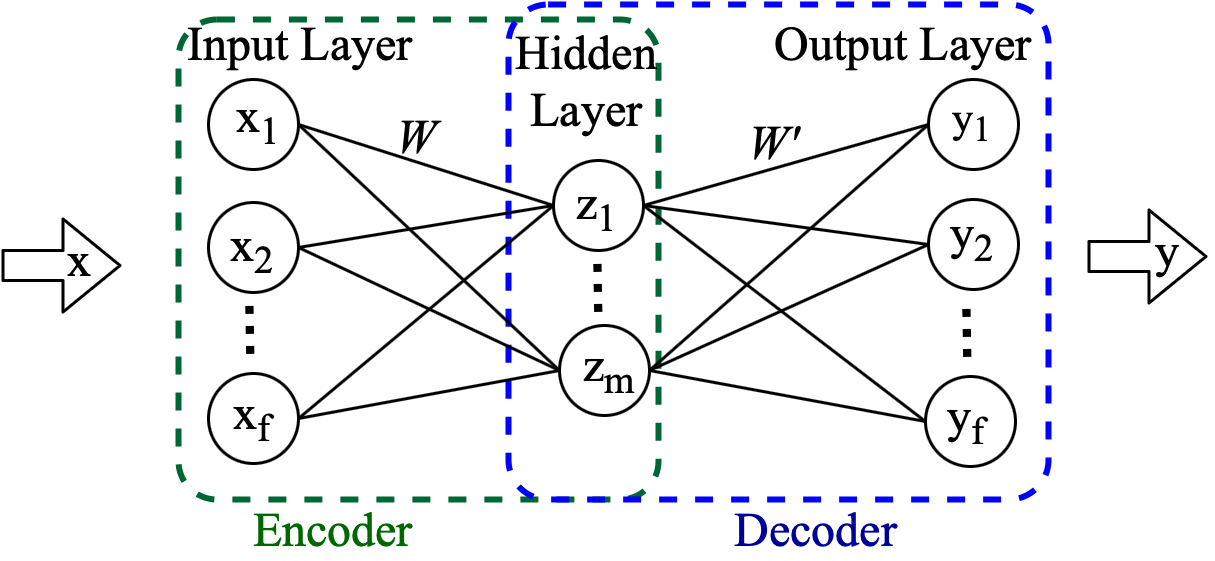}
\caption{Conventional Autoencoder structure.
\label{fig:AE}}
\end{figure}

As shown in Figure~\ref{fig:AE}, a~conventional autoencoder is a feed-forward neural network consisting of an input, an~output, and~one or more hidden layers. The~encoder part of the autoencoder reduces dimensionality. The~input $x$ of dimension $f$ is multiplied by the weights $W$ and, together with the bias $b$, is passed through the activation function  $\sigma$ to produce representation $z$ of dimension $m$, $m<f$ \cite{deeplearnbook}, as~follows.

% \vspace{-2pt}
\begin{equation}
z = \sigma(Wx + b)
\end{equation}
%   \vspace{-15pt}

Next, the~decoder attempts to reconstruct the input $x$ from the encoded value $z$. The~product of weights $W'$ and $z$ is added to biases $c$, and~the resultant is utilized as an input to the activation function $\sigma$ to generate the reconstructed signal $y$ as follows:

% \vspace{-2pt}
\begin{equation}
y = \sigma(W'z + c)
\end{equation}
% \vspace{-10pt}

Over a number of iterations (epochs), the~autoencoder optimizes the weights and biases by minimizing an objective function, such as  mean squared error (MSE) \cite{deeplearnbook}:
\begin{equation}
MSE = \frac{1}{f}\sum_{i=1}^{f}(x_i-y_i)^2
\end{equation}

\textls[-5]{The described feed-forward autoencoder ignores the spatial structure present in data and, to~address this issue, the~CAE was introduced~\cite{ConvAE}. A~CAE replaces the fully connected layers of the feed-forward autoencoder with convolutional and deconvolutional layers. The~convolutional layer performs the convolution operation on local spatial regions of the input. The~two-dimensional (2D) and one dimensional (1D) convolutions have led to significant advancements in image processing~\cite{CNN2} and sensor data processing~\cite{CNN1, cnn1_v2, Cnnv3}, respectively. In~this study, the~CAE-HIFD employs a 1D-CAE for HIF detection. In~addition to the convolutional layer, the~encoder typically has a max-pooling layer(s) which perform(s) down-sampling to reduce dimensionality. The~decoder in the CAE reconstructs the input with the help of transposed convolution or up-sampling layer(s).}

\section{Convolutional Autoencoder for HIF~Detection} \label{sec:methodology}

Traditionally, for~fault detection, autoencoders are trained with normal data and then used to detect abnormal operating conditions by identifying deviations from the learned normal data~\cite{araya2017ensemble}. On~the contrary, in~the proposed CAE-HIFD, the~CAE is trained using fault data only and recognizes non-fault operating conditions by detecting deviations from the learned fault scenarios. The~spatial feature learning and generalization capability of the CAE assist the CAE-HIFD to detect new HIF scenarios that are not present in the training set. Furthermore, since the CAE is only trained on the fault data, any non-fault cases will not be identified as HIF, which increases the security of the proposed protection~strategy.

\begin{figure}[H]
%\centering
\widefigure
\includegraphics[width=16 cm]{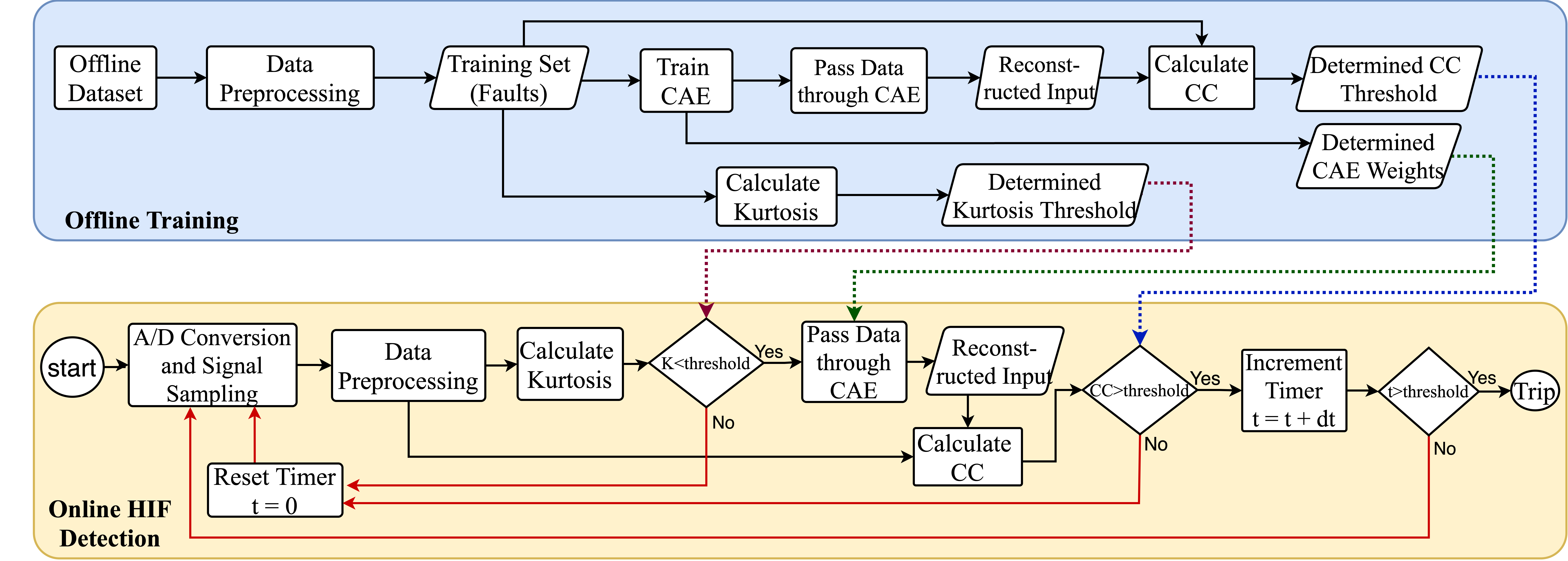}
\caption{CAE Framework for HIF Detection (CAE-HIFD).\label{fig:framework}}
\end{figure}

As depicted in Figure~\ref{fig:framework}, the~CAE-HIFD is comprised of offline training and online HIF detection. The~analog three-phase voltage and current signals are sampled and converted to digital signals using A/D converters. Training happens in the offline mode using a dataset prepared from multivariate time series consisting of three-phase voltage and current signals. The~online HIF detection uses the weights and thresholds obtained from the offline training. As~the data preprocessing step is the same for both training and detection, the~preprocessing is described first, followed by the explanation of training and~detection.

\end{paracol}

\begin{paracol}{2}
%\linenumbers
\switchcolumn
\vspace{-10pt}

\subsection{Data~Preprocessing}

The first step of data preprocessing entails the sliding window approach applied to time-series data to transform it into a representation suitable for the CAE. As~illustrated in Figure~\ref{fig:slidingW}, the~first $n$ samples (readings) make the first data window; thus, the~data window dimension is $n$ time steps $\times$  $f$  number of features. In~each iteration of the algorithm, the~data window slides for $s$ time steps, where $s$ is referred to as a \textit{stride}, to~create the second data window and so on. Note that Figure~\ref{fig:slidingW} illustrates a case where $s$ = $n$.

\begin{figure}[H]
%\centering
\includegraphics[width=10.5 cm]{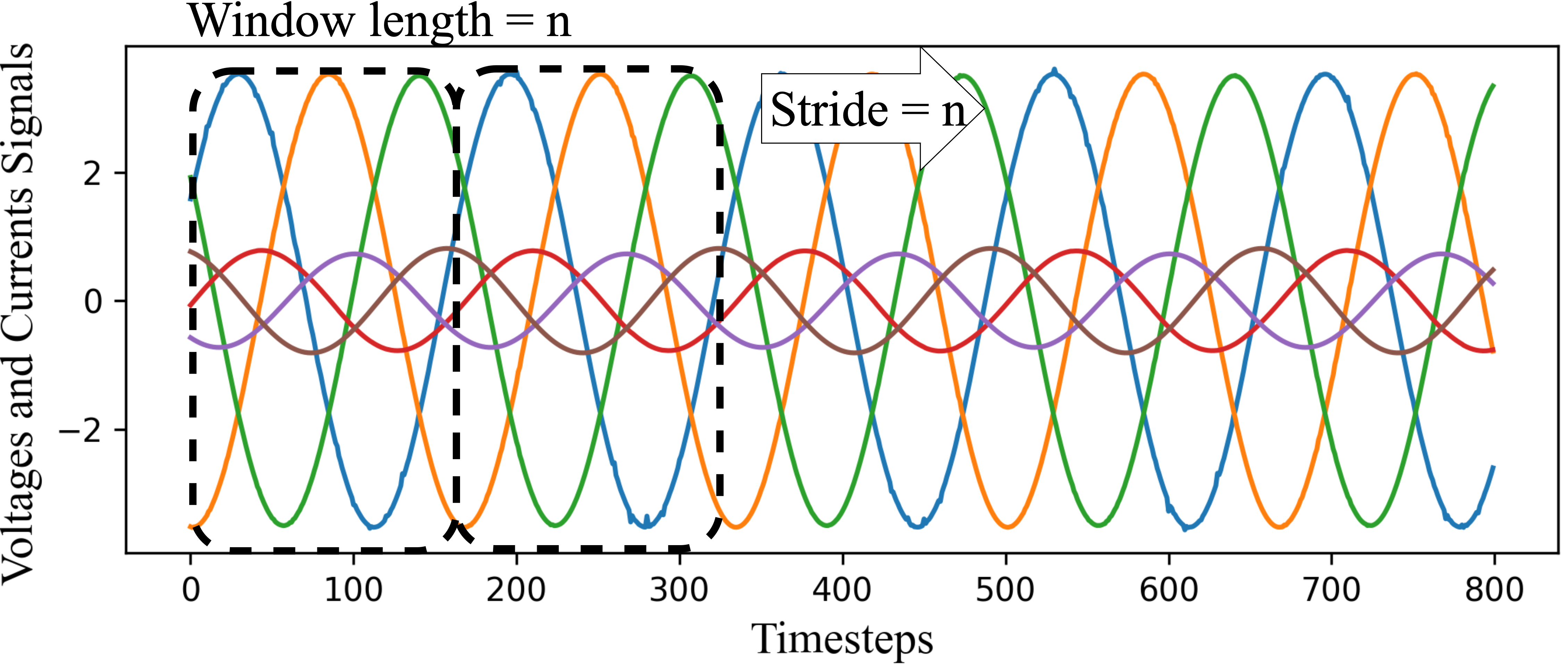}
\caption{Sliding window~approach.\label{fig:slidingW}}
\end{figure}

Each voltage and current phase signal (each feature) in the sliding window is processed individually by differencing. The~first-order $d^1$ and second-order $d^2$ differencing of signal $y(t)$, e.g.,~phase A voltage, are as follows:
% \vspace{-5pt}
\begin{equation}
d^1(t)=y(t)-y(t-1)
\end{equation}
\begin{equation}
d^2(t)=d^1(t)-d^1(t-1)
\end{equation}

The HIF causes small distortions in the voltage and current waveforms. The~second-order differencing helps the CAE to learn and detect the HIF pattern by amplifying these distortions and suppressing the fundamental frequency component of each input signal. Differencing also amplifies noise; nevertheless, the~generalization and spatial feature extraction capabilities of the CAE make the CAE-HIFD robust against noise as demonstrated in Section~\ref{sec:results}.

\subsection{Offline~Training}

The CAE is trained solely with the fault data, and~the non-fault data are only utilized for the system validation. As~illustrated in Figure~\ref{fig:framework}, the~preprocessed data are passed to the CAE as one data window, $n \times f$ matrix, at~a time. As~shown in Figure~\ref{fig:CAE}, the~CAE is composed of two main components: encoder and decoder~\cite{deeplearnbook}.

\begin{figure}[H]
%\centering
\includegraphics[width=12 cm]{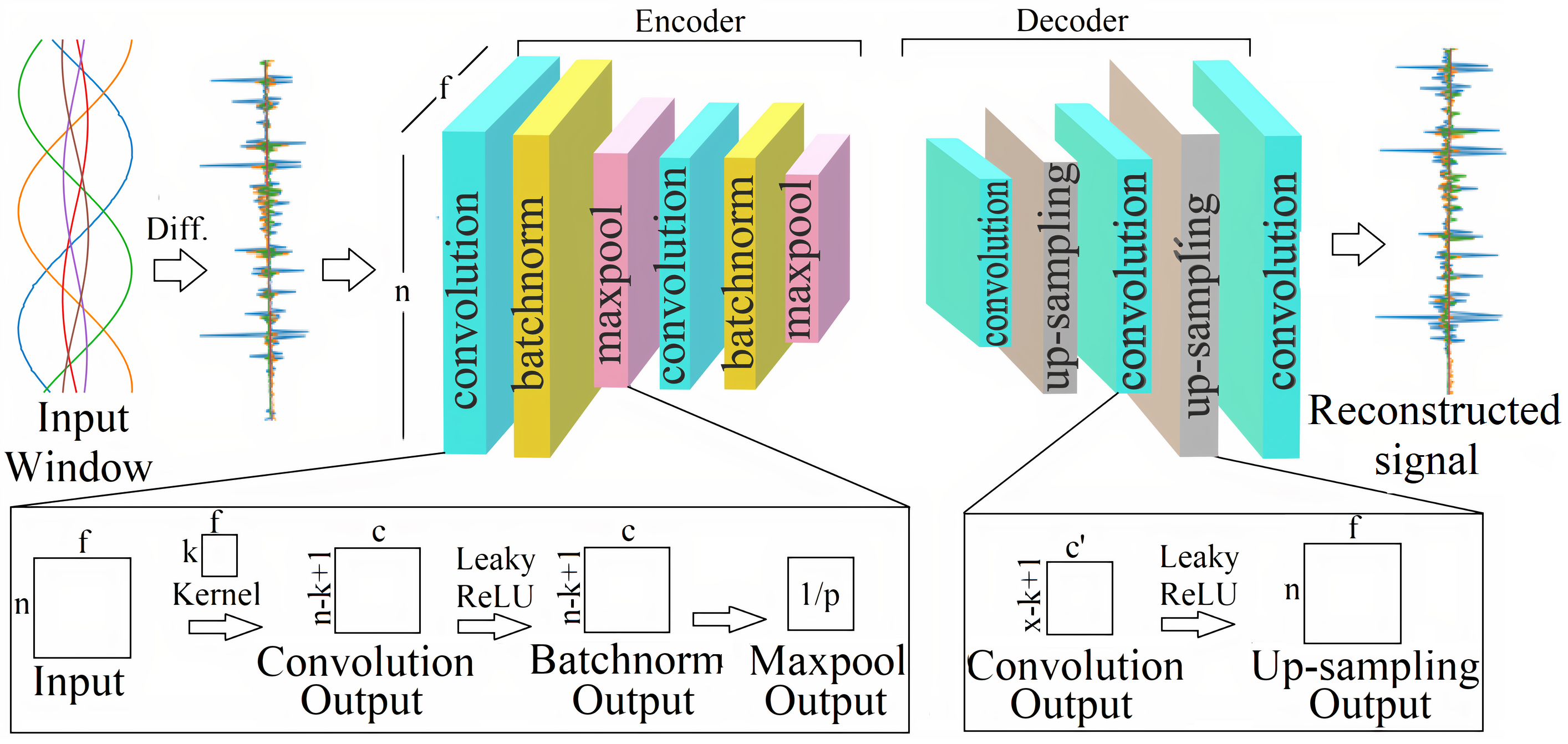}
\caption{Convolutional autoencoder (CAE) structure.\label{fig:CAE}}
\end{figure}

The first layer in the encoder performs the 1D convolution operation on the $n\times f$ input matrix with the kernel of size \textit{k} $\times$ \textit{f}. This kernel moves across the time steps of the input and interacts with $k$ time steps (here $k<n$) of the input window at a time; thus, during~the CAE training, the~kernel learns the local spatial correlations in the input samples. There are $m$ kernels in the first layer and each kernel convolves with input to generate an activation map. Consequently, the~output of the first layer has a dimension of $(n-k+1) \times m$, and~every column of this output matrix corresponds to the weights of one kernel. These kernel weights are learned during the CAE training process. Rectified linear unit (ReLU) activation function is often used to introduce non-linearity after the convolution. However, here LeakyReLU, a~leaky version of ReLU, is used instead because ReLU discards the negative values in the sinusoidal wave~\cite{deeplearnbook}. Next, the~batch normalization layer re-scales and re-centers data before passing them to the next layer in order to improve the training convergence. The~batch normalized data are passed to the max-pooling layer to reduce data dimensionality and the associated computational complexity. The~size of the max-pooling operation is $p$; therefore, the~output of the pooling layer is $\frac{1}{p}$ of the convolved input. As~illustrated in Figure~\ref{fig:CAE}, the~convolution, batch normalization, and~max-pooling layers are repeated two times to extract features on different levels of abstraction. These encoder layers create an encoded representation of the input signal which is passed to the~decoder.

Although the encoder decreases the dimensionality of the input, the~decoder reconstructs the original signal from these encoded values. In~the decoder, as~illustrated in \mbox{Figure~\ref{fig:CAE}}, the~convolutional layer first generates the activation map, and~then the up-sampling operations increase the dimensionality of the down-sampled feature map to the input vector size. During~up-sampling, the~dimensionality of the input is scaled by repeating every value along the time steps in the signal with the scaling factor set according to the max-pooling layer size in the encoder. Similar to the encoder, in~the decoder, the~convolutional and up-sampling layers are repeated twice (Figure \ref{fig:CAE}).

The CAE optimizes the weights and the biases using the back-propagation process in which the gradient descent is applied based on the loss function, typically MSE. In~the proposed CAE-HIFD, the~MSE is utilized as the loss function for training the CAE using fault data. In~an autoencoder, the~MSE is also referred to as a reconstruction error as it evaluates the similarity between the input signal and the reconstructed signal given by the autoencoder output. As~the objective of the gradient descent algorithm is to minimize the MSE for training data, the~MSE is expected to be low for the training data and high for any deviations from the training~patterns.

In the CAE-HIFD, the~CAE sees only the fault data during training, and~consequently, the~trained CAE is expected to fail in reconstructing the non-fault data input. Therefore, the~MSE for the non-fault data is expected to be higher than the MSE for the learned fault data. Traditionally, in~autoencoders, the~separation between fault and non-fault data is done based on a threshold which is determined using the MSEs of the training dataset. However, in~HIF detection, when CAE is trained with  fault data, MSE is not a reliable metric for calculating the threshold. As~illustrated in Figure~\ref{fig:corr}, the~differentiated fault data forms a complex pattern with a high number of fluctuations causing the dissimilarities between the CAE output and input. The~magnitude of these fluctuations varies from $-$2.0 to 1.0 and, as~a result, even a small mismatch between input and CAE output leads to high MSE: for example, in~Figure~\ref{fig:corr}a, MSE for fault data window is 0.0244. On~the other hand, in~Figure~\ref{fig:corr}b, MSE for steady state data window is 0.0002 because of a relatively simpler pattern compared to HIFs and small amplitudes of differentiated signal oscillations varying from $-$0.04 to 0.04. Consequently, the~MSE is not a reliable indicator to discriminate between HIF and non-fault~cases.

In signal processing, a~metric commonly used to evaluate the similarity between signals is the cross-correlation (CC) \cite{signalstats} which is defined as:
% \vspace{-2pt}
\begin{equation}
CC=(f*g)(\tau) = \int_{-\infty}^{\infty} \overline{f(t)}g(t+\tau) dt
\end{equation}

% just changed CORR to CC
\noindent where $f$ and $g$ are two signals and $\tau$ is a time shift in the~signal.

\begin{figure}[H]
%\centering
\includegraphics[width=10.5 cm]{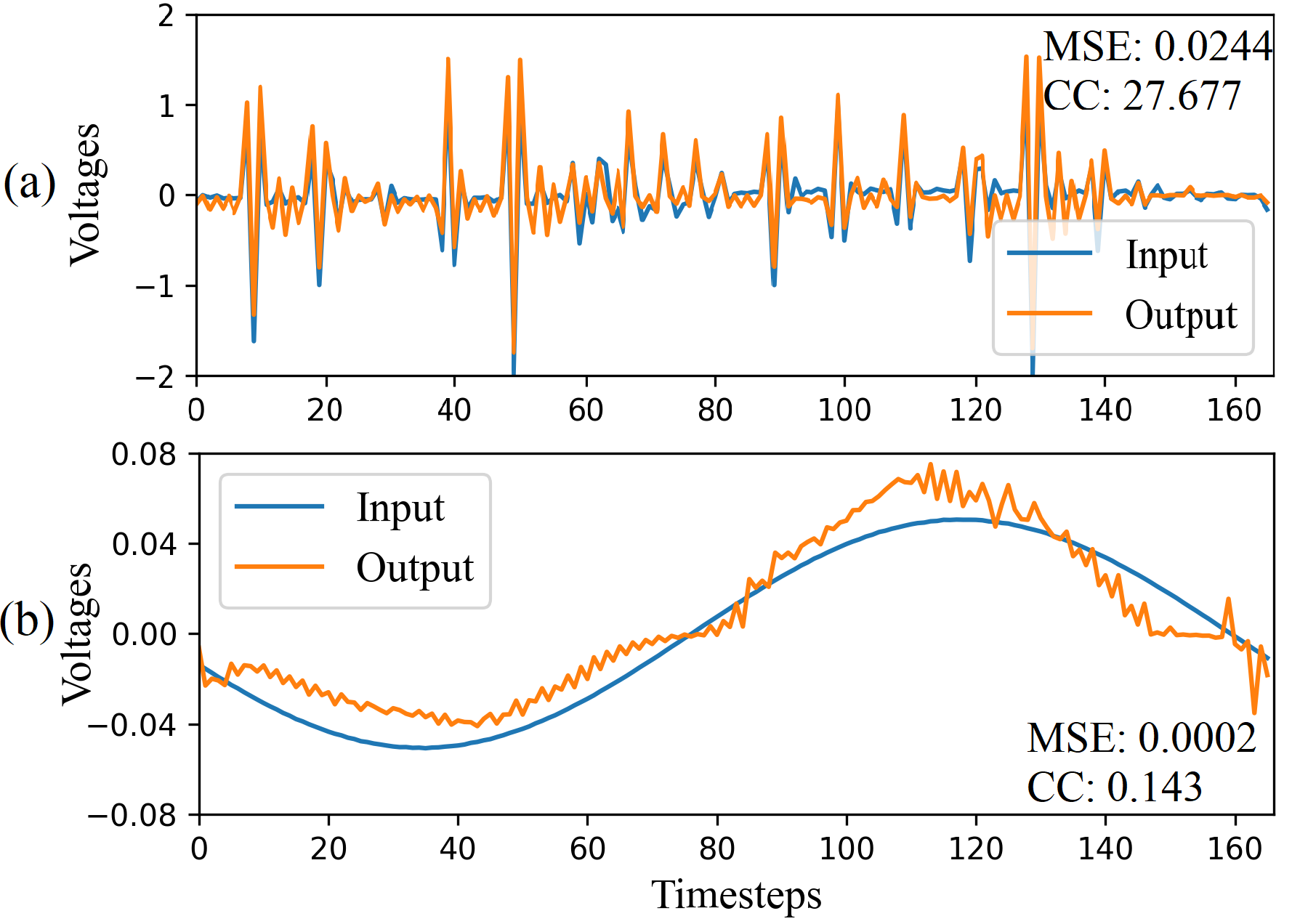}
\caption{Input and output of the CAE for (\textbf{a}) HIF and (\textbf{b}) Non-HIF~scenario.\label{fig:corr}}
\end{figure}

In CAE-HIFD, CC is used to measure the similarity between the CAE input and output signals. As~illustrated in Figure~\ref{fig:framework}, after~the CAE training is completed, the~trained CAE reconstructs all data windows from the training set and obtains reconstructed signals. Next, for~each window in the training set, the~CC value is calculated for the input signal and the corresponding CAE output. As~seen in Figure~\ref{fig:corr}a, the~HIF data window has a CC value of 27.677 because the input and output signals of the CAE are similar. On~the contrary, a~normal steady-state operating condition data window, Figure~\ref{fig:corr}b, has a low CC value of 0.143 as the output deviates from the input. As~the minimum CC value from the training set represents the least similar input--output pair from the training set, this minimum CC value serves as the CC threshold for separating HIF and non-HIF~cases.

The CAE perceives the responses to disturbances, such as capacitor and load switching, to be HIFs because these disturbances cause waveform distortions. However, these disturbances usually occur for a shorter duration of time than HIFs making their statistical distribution different from those of HIFs and steady-state operation. Figure~\ref{fig:K} shows that the disturbances and HIFs both exhibit Gaussian behavior, but~the disturbances have a thinner peak and flatter tails on the probability density function (PDF) plot. In~contrast, steady-state operation data (sinusoidal waveforms) have an arcsine~distribution.

To distinguish disturbances from HIFs, the~statistical metrics kurtosis is used. The~kurtosis provides information about the tailedness of the distribution relative to the Gaussian distribution~\cite{signalstats}. For~univariate data $y_1,y_2,y_3,…,y_n$ with standard deviation $s$ and mean $\bar{y}$, the~kurtosis is: %defined as:
\begin{equation}
K = \frac{\sum_{i=1}^{n} (y_i - \bar{y})^4/n}{s^4}
\end{equation}

As Figure~\ref{fig:K} shows, flatter tails and thinner peaks results in higher kurtosis values. For~example, the~distribution of the differentiated capacitor switching disturbance in Figure~\ref{fig:K}b has a kurtosis value of $K=76.6$ which is higher than the $K=1.9$ for the HIF distribution in Figure~\ref{fig:K}f.

The kurtosis is calculated from the training set individually for each data window after applying differencing. To~prevent misinterpretation of the $K$ values and avoid treating HIFs as non-fault disturbances, the~kurtosis threshold must be higher than every $K$ value present in the training set. Accordingly, the~kurtosis threshold is the value below which all the $K$ values of the training data~lie.

\begin{figure}[H]
%\centering
\includegraphics[width=10.5 cm]{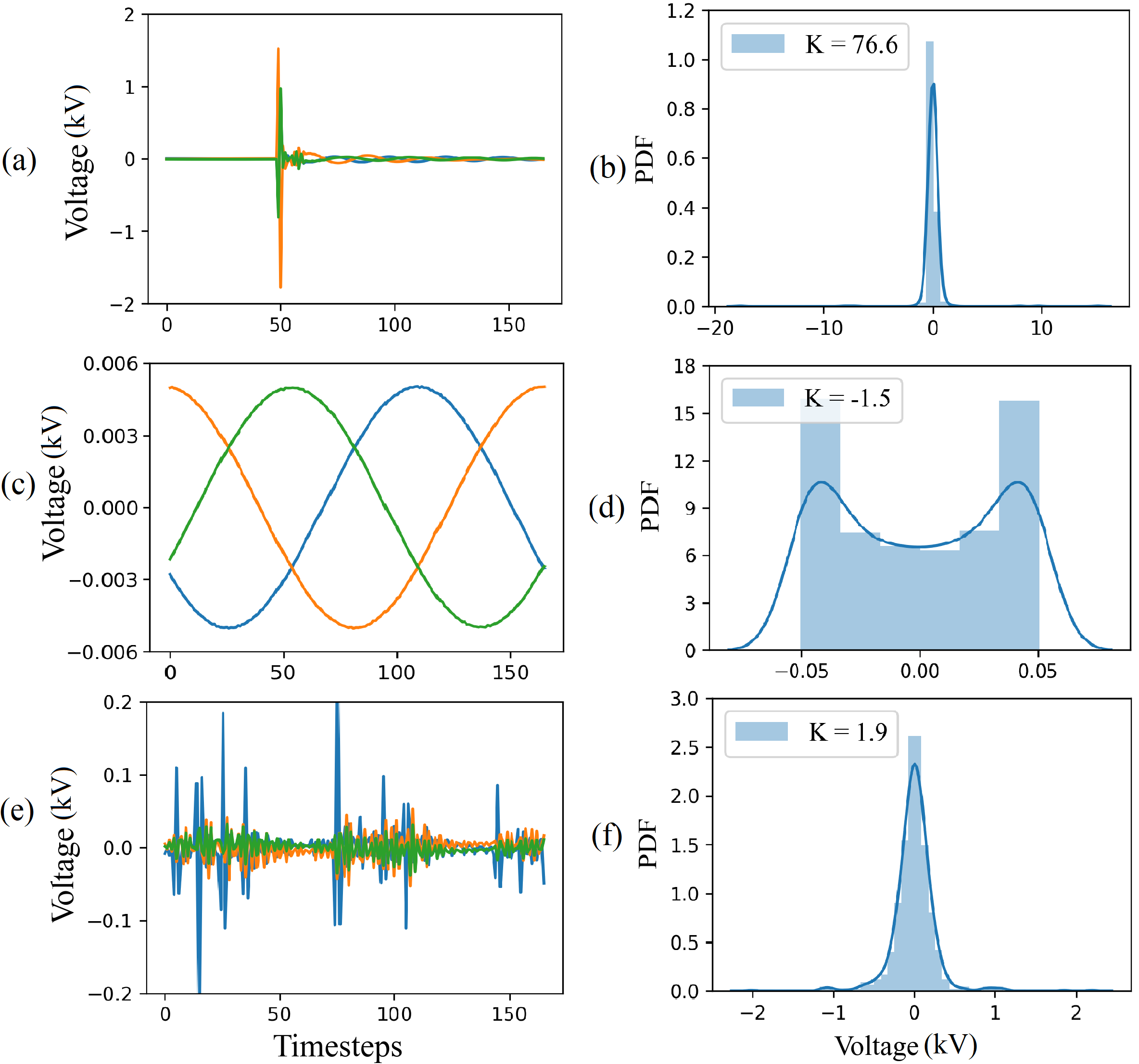}
\caption{Kurtosis analysis: (\textbf{a}) differentiated voltage signal corresponding to a capacitor switching disturbance, (\textbf{b}) PDF of the voltage signal corresponding to a capacitor switching disturbance (\textbf{c}) differentiated normal steady-state voltage, (\textbf{d}) PDF of the normal steady-state voltage, (\textbf{e}) differentiated HIF voltage, and~(\textbf{f}) PDF of the HIF~voltage.\label{fig:K}}
\end{figure}

The artifacts of the offline training are the CC threshold, the~learned CAE weights, and~the kurtosis threshold. These artifacts are used for online HIF~detection.

\subsection{HIF~Detection}

The online HIF detection algorithm uses the artifacts generated by offline training as illustrated in Figure~\ref{fig:framework}. First, the~analog input signal is converted to digital by the A/D converter and the data preprocessing module generates data windows which proceed through the remaining HIF detection components, one window at the~time.

The value of kurtosis is calculated for each data window and compared with the corresponding threshold obtained from the offline training. Any data window with the kurtosis value above the threshold is identified as a non-fault disturbance case for which the CAE is disabled because there is no need for additional processing as the signal is already deemed to be a disturbance. Next, the~timer is reset for processing the next input signal~segment.

If the kurtosis value is less than the threshold, the~data window is sent to the trained CAE which encodes and reconstructs the signal. As~the CAE is trained with fault data, for~HIFs, the~reconstructed signal is similar to the original signal. This similarity is evaluated by calculating the CC between the reconstructed signal and the original signal. If~the CC value of the data window is greater than the CC threshold determined in the training process, the~signal is identified to be corresponding to a~HIF.

Under transient disturbances, such as capacitor switching, the~value of CC may exceed the corresponding threshold for a short time period immediately after the inception of disturbance. False identification of disturbances as HIFs is prevented using a pick-up timer. The~timer is incremented when the CC exceeds its threshold and is reset to zero whenever the CC or K indicates a non-HIF condition, as~shown in Figure~\ref{fig:framework}. A~tripping (HIF detection) signal is issued when the timer indicates that the time duration of the HIF exceeds a predetermined~threshold.

\section{Evaluation} \label{sec:results}

This section first describes the study system and the process of obtaining data for the performance verification studies. Next, the~details of CAE-HIFD model training and the effects of different CAE-HIFD components are presented. Furthermore, the~response of the CAE-HIFD to different case studies is demonstrated. Finally, the~CAE-HIFD performance is compared with other HIF detection approaches, and~its sensitivity to noise is~examined.

% \vspace{-7pt}

\subsection{Study~System} \label{sec:studies}

The dataset utilized for model training and evaluation is obtained through time-domain simulation studies performed in the PSCAD software. The~study system is the IEEE 13 node test feeder of Figure~\ref{fig:IEEE13n}, a~realistic 4.16 kV distribution system with a significant load unbalance. This test feeder was selected in order to examine the system behavior under challenging load unbalance conditions and because of its common use in HIF studies~\cite{pow17}. Detailed information regarding the line and the load data are provided in the Appendix~\ref{app:a}, and~further information about this benchmark system can be found in~\cite{pow17}. For~an accurate representation of the HIF behavior, the~antiparallel diode model of Figure~\ref{fig:nonlinHIF}, \cite{pow2, pow3, pow4, pow7, pow8} is utilized. The~HIF model parameters representing seven different faulted surface types are given in Table~\ref{tab3} \cite{pow2,pow9}. These parameters lead to effective fault impedances as high as 208 ohms in 4.16 kV distribution system.

\begin{figure}[H]
%\centering
\includegraphics[width=10.5 cm]{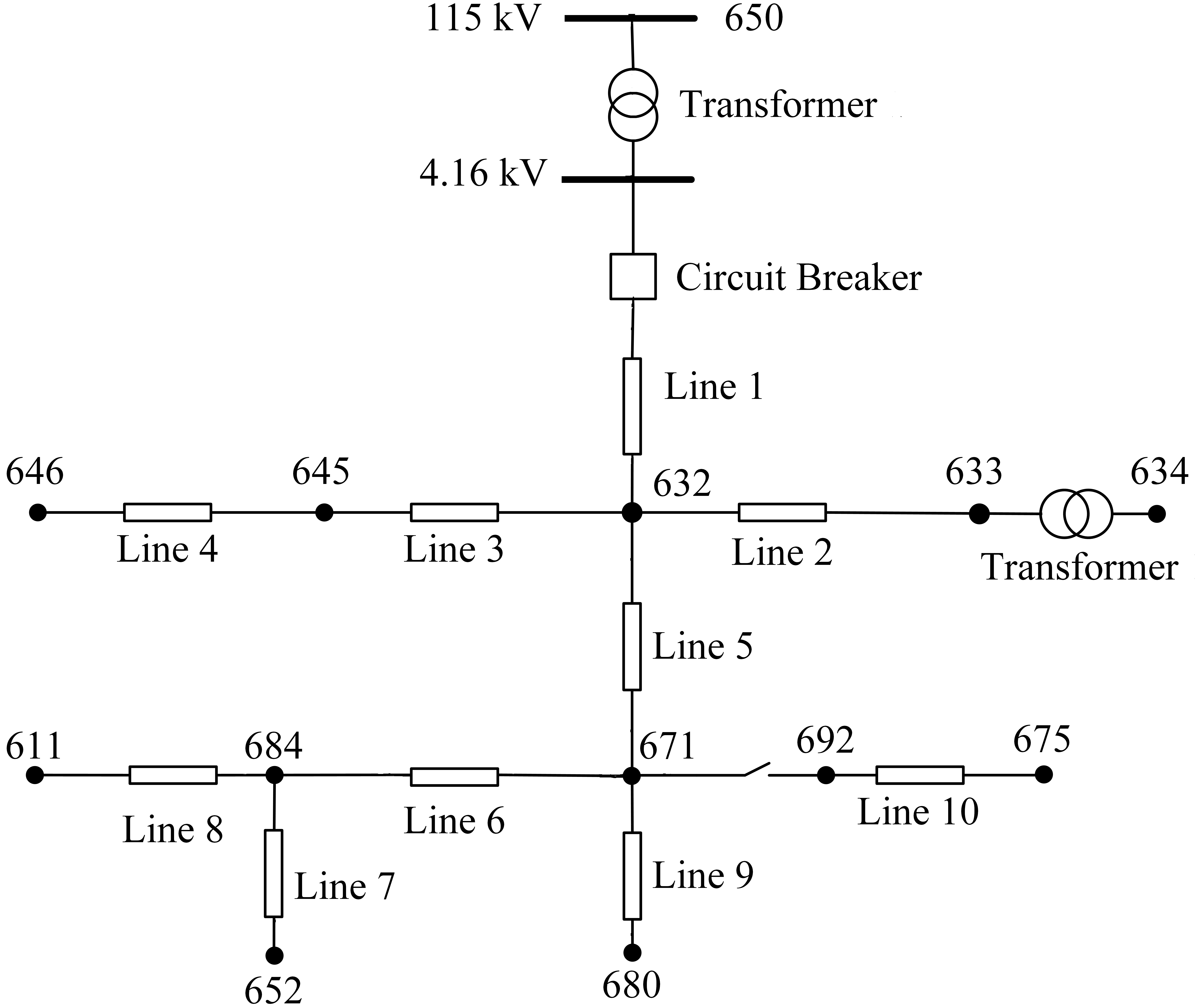}
\caption{IEEE 13 node test~feeder.\label{fig:IEEE13n}}
\end{figure}
\unskip

\begin{figure}[H]
%\centering
\includegraphics[width=2.5 cm]{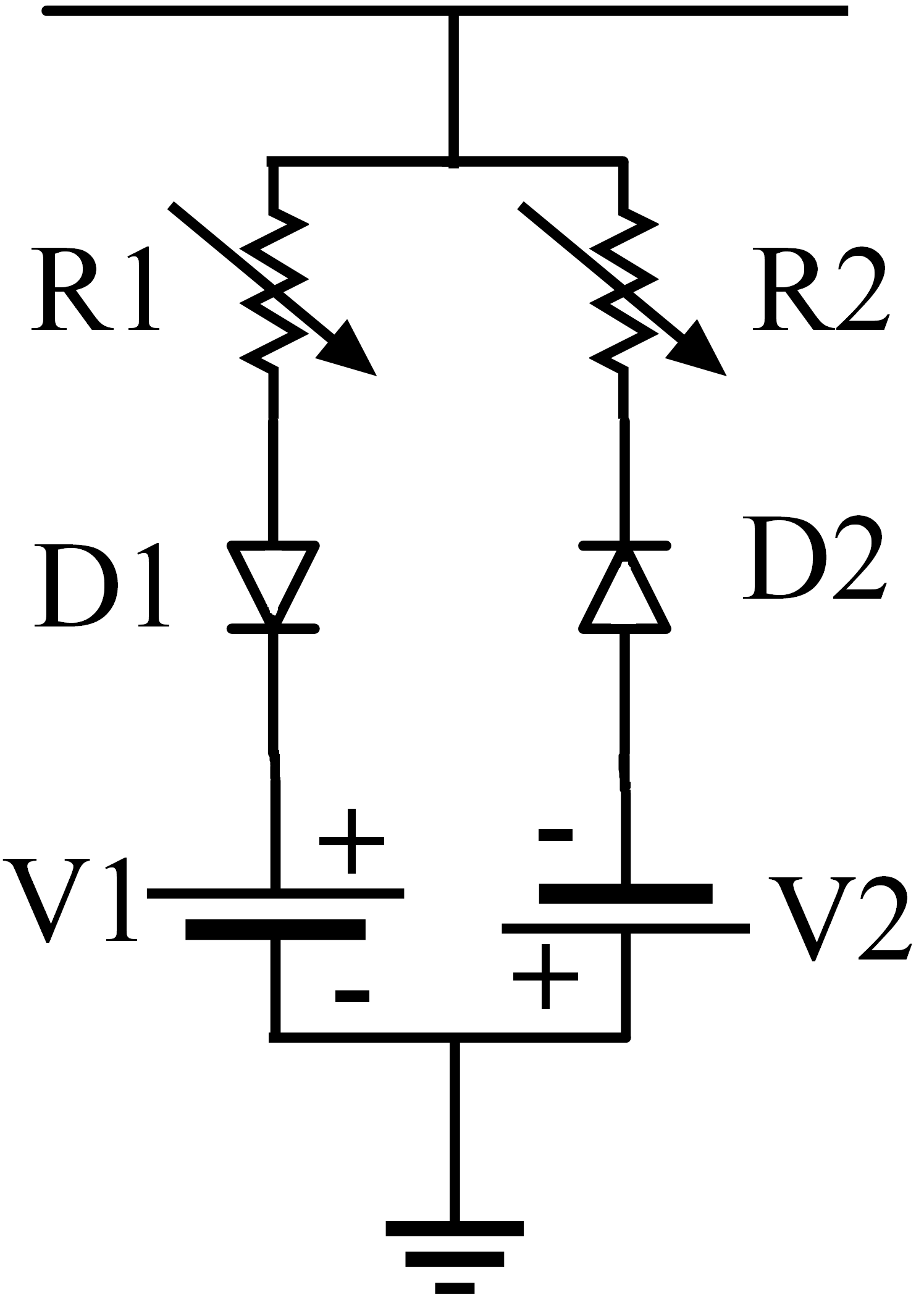}
\caption{Nonlinear HIF model utilized for time-domain simulation~studies.\label{fig:nonlinHIF}}
\end{figure}

In total, 210 faulty cases were simulated: 7 different surfaces, 10 fault locations, and~3 phases. After~the windowing technique, this resulted in 1372 HIF data windows. Additionally, the~dataset obtained from simulations contained 272 non-fault data windows. Using data obtained from simulation studies enables considering diverse fault types, locations, and~surfaces while obtaining such diverse data from real-world experiments would be difficult or even impossible. Of the fault data, 80\% are assigned for the model training and the rest for testing. As~CAE-HIFD requires only fault data for training, all non-fault data are assigned for testing. Of the training set, 10\% are used as a validation set for the hyperparameter optimization.
% \vspace{-4pt}

% The MDPI table float is called specialtable
\begin{specialtable}[H]
\setlength{\tabcolsep}{4.85mm}
%\centering
\caption{HIF Model parameters For different~surfaces.\label{tab3}}
%%% \tablesize{} %% You can specify the fontsize here, e.g.,~\tablesize{\footnotesize}. If commented out \small will be used.
\begin{tabular}{ccccc}
\toprule
\textbf{Surfaces} & \textbf{R1 (\boldmath$\Omega$)} & \textbf{R2 (\boldmath$\Omega$)} & \textbf{V1 (\boldmath$V$)} & \textbf{V2 (\boldmath$V$)}\\

\midrule
Wet Sand & 138 $\pm$ 10\% & 138 $\pm$ 10\% & 900 $\pm$ 150 & 750 $\pm$ 150 \\
Tree Branch & 125 $\pm$ 20\% & 125 $\pm$ 20\% & 1000 $\pm$ 100 & 500 $\pm$ 50 \\
Dry Sod & 98 $\pm$ 10\% & 98 $\pm$ 10\% & 1175 $\pm$ 175 & 1000 $\pm$ 175 \\
Dry Grass & 70 $\pm$ 10\% & 70 $\pm$ 10\% & 1400 $\pm$ 200 & 1200 $\pm$ 200 \\
Wet Sod & 43 $\pm$ 10\% & 43 $\pm$ 10\% & 1550 $\pm$ 250 & 1300 $\pm$ 250 \\
Wet Grass & 33 $\pm$ 10\% & 33 $\pm$ 10\% & 1750 $\pm$ 350 & 1400 $\pm$ 350 \\
Rein. Concrete & 23 $\pm$ 10\% & 23 $\pm$ 10\% & 2000 $\pm$ 500 & 1500 $\pm$ 500 \\

\bottomrule
\end{tabular}
\end{specialtable}

\subsection{CAE-HIFD Model~Training}
The performance of the CAE is evaluated using accuracy (Acc), security (Sec), dependability (Dep), safety (Saf), and~sensibility (Sen) \cite{formulas}.
\vspace{-3pt}
\begin{align}
Acc &= \frac{TP+TN}{TP+TN+FP+FN} \times 100 \% \\
Sec &= \frac{TN}{TN+FP} \times 100 \% \\
Dep &= \frac{TP}{TP+FN}  \times 100 \% \\
Saf &= \frac{TN}{TN+FN}  \times 100 \% \\
Sen &= \frac{TP}{TP+FP}  \times 100 \%
\end{align}

Here, true positives (TP) and true negatives (TN) are the numbers of correctly identified fault and non-fault cases, and~false negatives (FN) and false positives (FP) are the numbers of miss-classified fault and non-fault cases. The~accuracy is the percentage of overall correctly identified states, the~security is the healthy state detection precision, the~dependability is the fault state detection precision, the~safety is resistance to faulty tripping, and~the sensibility is resistance to unidentified faults~\cite{formulas}. Note that dependability is also referred to as true positive rate (TPR) or sensitivity, while security is referred to true negative rate (TNR) or specificity {\cite{araya2017ensemble}}.

To achieve high accuracy, the~CAE hyperparameters must be tuned; this includes the number and size of kernels, learning rate, optimizer, and~batch size. Hyperparameter tuning is performed with grid search cross-validation (GSCV), wherein an exhaustive search is conducted over pre-specified parameter ranges. The~GSCV determines the best performing parameters based on the scoring criteria provided by the model, in~our case accuracy. The~tuned CAE has 256-128-128-256 filters of size $3 \times 3$ in the four convolution layers, the~optimizer is Adam, the~learning rate is 0.001, and~the batch size is~16.

The window size after differencing is 166 voltage/current samples, which corresponds to one cycle of the 60 Hz power frequency signal sampled at a rate of 10 kHz. The~proposed method does not operate based on the fundamental frequency components of the input signal and, thus, is not sensitive to frequency deviations as shown in Section \ref{subsec:freq}. The~sliding window stride during training impacts the CAE-HIFD performance: its value is determined on the training dataset. Once the system is trained, it is used with the stride of one. The~upper bound for the stride value is 166 as a  higher value would lead to skipped samples. The~performance metrics for varying stride values are shown in Figure~\ref{fig:strivar}. It can be observed that the safety and dependability are not affected by the change in the stride value as none of the non-HIF cases are misclassified as a HIF case. The~accuracy, security, and~sensibility are 100\% for the stride size of 166, and~for shorter strides, these metrics are slightly lower. As~the stride decreases, a~few data windows are mistakenly identified as faults resulting in decrease in security, sensibility, and~accuracy. Hence, the~stride value of 166 is selected for the sliding window in the preprocessing of the training dataset.

\begin{figure}[H]
%\centering
\includegraphics[width=10.5 cm]{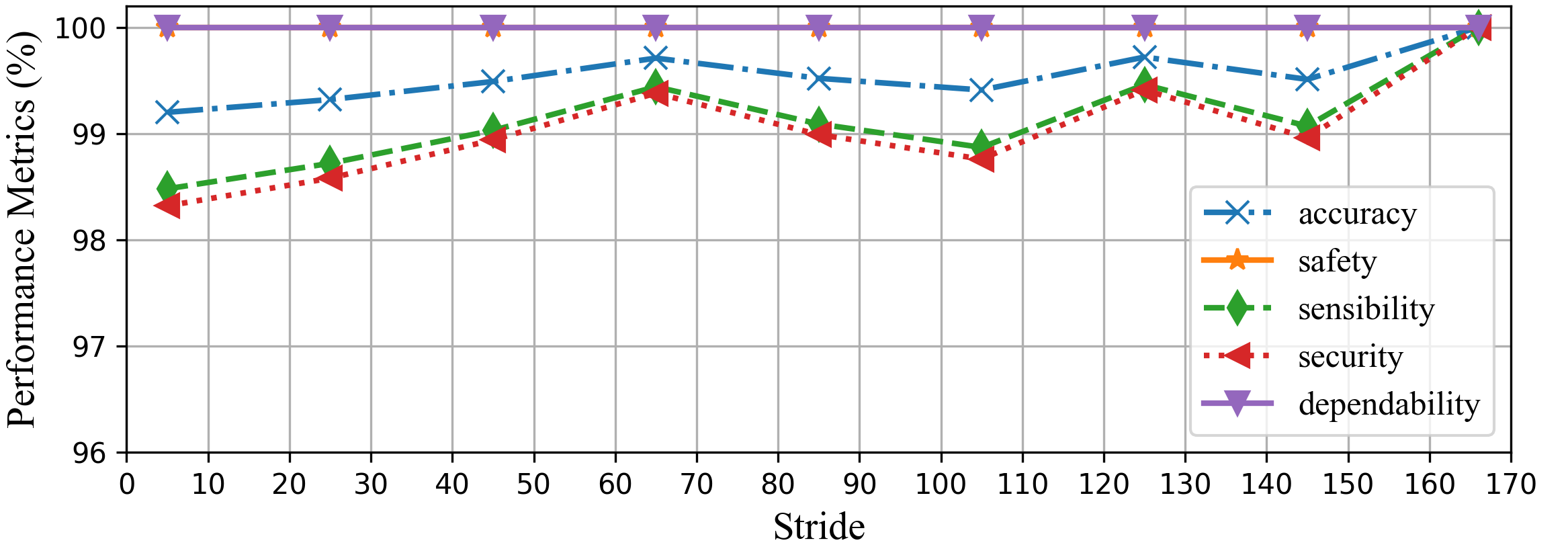}
\caption{Impact of the sliding window stride on the performance metrics of the~CAE-HIFD.\label{fig:strivar}}
\end{figure}

The CAE-HIFD prevents false detection of disturbances as HIFs using kurtosis value. As~the distributions of the HIFs and disturbances both exhibit Gaussian behavior, an~HIF window can have a $K$ value close to the $K$ value of a disturbance. The~kurtosis threshold is determined starting from the kurtosis values for which all HIFs scenarios in the training data lie below this threshold: in this case 10. Next, accuracy on the training data is examined with thresholds close to this initial threshold. As~illustrated in Figure~\ref{fig:kurtvar}, the~accuracy is 100\% when kurtosis thresholds are between 9.5 and 10.5. The~accuracy decreases when threshold is below 9.5 because some of the HIF scenarios are mistakenly detected as non-HIF scenarios ($FN>0$). Furthermore, the~threshold above 10.5 leads to low accuracy as some non-HIF scenarios are falsely declared as HIFs  ($FP>0$). Consequently, the~kurtosis threshold of 10 is selected to discriminate the non-fault disturbances from the~HIFs.

\begin{figure}[H]
%\centering
\includegraphics[width=10.5 cm]{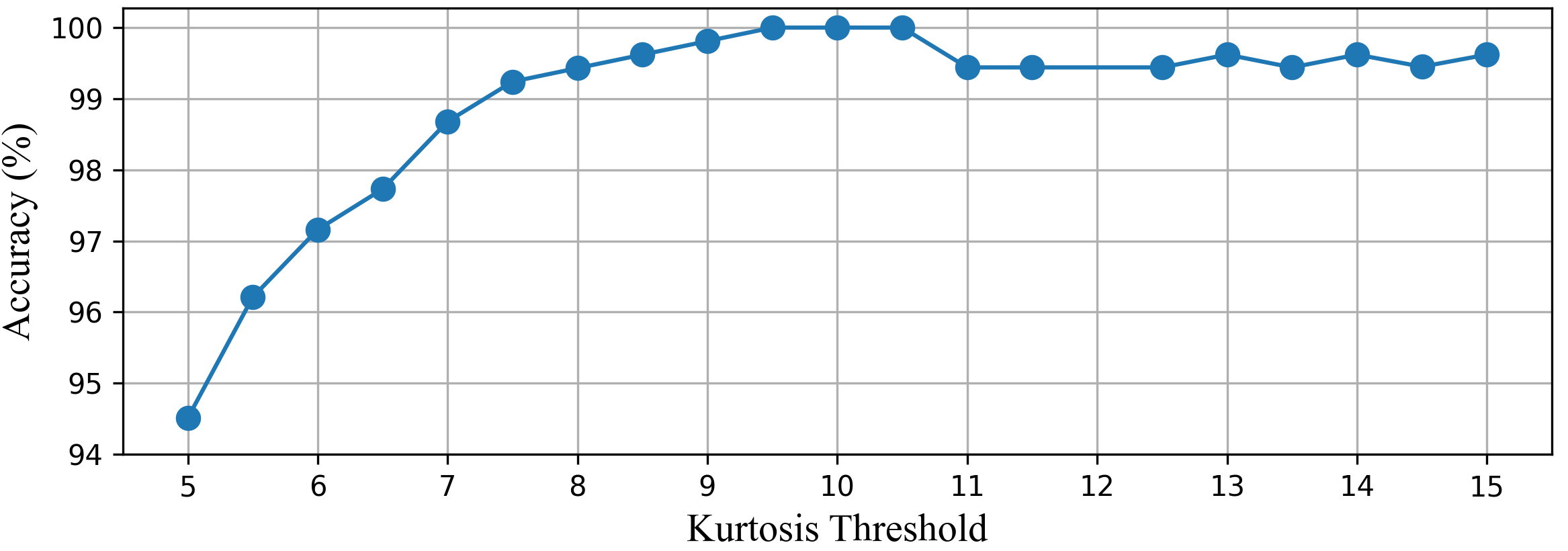}
\caption{Impact of the kurtosis threshold on the accuracy of the~CAE-HIFD.\label{fig:kurtvar}}
\end{figure}
\unskip

\subsection{Effects of CAE-HIFD's~Components}

The proposed CAE-HIFD uses differencing and cross-correlation in addition to the main component, the~CAE, to~increase various performance metrics. Furthermore, kurtosis is utilized to improve the security of the proposed method. Consequently, as~depicted in Table~\ref{tab2}, the~CAE-HIFD achieves 100\% performance in all five considered metrics regardless of the surface type, inception angle, and~fault~locations.

Additionally, Table~\ref{tab2} includes variants of the CAE-HIFD with only some of the three components included. With~only CC and kurtosis, the~accuracy and sensibility drop to nearly 51\%, and~the security decreases by 99.6\%. In~the absence of differencing, the~CAE cannot learn patterns to distinguish between the HIF and the non-HIF data windows and, as~a result, a~large number of the non-HIF data windows are falsely classified as the HIF which means high FP, thus low~security.

To examine the impact of CC, the~traditional MSE is used in place of CC to measure the similarity of the input and reconstructed signal. As~shown in Table~\ref{tab2}, in~the absence of CC, the~values of accuracy and sensibility drop to nearly 50\%. This happens because the MSEs calculated for the HIF and non-HIF data windows are similar and, thus, non-HIFs are falsely detected as HIFs. Furthermore, the~security value is low (0.40\%), whereas the dependability value is high (92.67\%) as there are only a few TN and FN compared to TP and~FP.

% The MDPI table float is called specialtable
\begin{specialtable}[H]
\setlength{\tabcolsep}{5.45mm}
%\centering
\caption{Impact of differencing, cross-correlation, and~kurtosis on CAE-HIFD~performance.\label{tab2}}
%%% \tablesize{} %% You can specify the fontsize here, e.g.,~\tablesize{\footnotesize}. If commented out \small will be used.
\begin{tabular}{cccccc}
\toprule
\textbf{Model} & \textbf{Acc} & \textbf{Saf} & \textbf{Sen} & \textbf{Sec} & \textbf{Dep} \\

\midrule
Proposed CAE-HIFD & 100 & 100 & 100 & 100 & 100 \\
CAE with CC, K & 51.19 & 100 & 51.10 & 0.37 & 100 \\
CAE with K, Diff & 48.29 & 4.76 & 50.10 & 0.40 & 92.67 \\
CAE with CC, Diff & 96.38 & 100 & 93.31 & 92.70 & 100 \\
CAE with CC & 51.10 & 100 & 50.64 & 1.84 & 100 \\
CAE with Diff & 48.9 & 37.93 & 49.51 & 4.04 & 93.43 \\
CAE with K & 89.64 & 96.67 & 84.16 & 82.77 & 96.35 \\

\bottomrule
\end{tabular}
\end{specialtable}

Omitting the kurtosis evaluation results in only a small increase in the number of FP cases which are the non-fault disturbances falsely declared as HIFs. Therefore, as~shown in Table~\ref{tab2}, all the performance metrics values decrease by less than 8\%.

Finally, only one out of the three components is included in the CAE-HIFD framework. Whereas the simultaneous use of differencing and cross-correlation achieves relatively high performance metrics, with~only one of the two components, there is a major decrease in security (more than 95\%). With~kurtosis only, all metrics are between 82\% and 97\% in comparison to 100\% obtained in the presence of all three components. This is caused by the absence of differencing which assists in amplifying sign wave distortions and omission of cross-correlation which facilitates signal comparisons. The~results shown in Table~\ref{tab2} highlight the necessity of each CAE-HIFD component and the contribution of each component to the HIF detection~performance.

\subsection{CAE-HIFD Response to Different Case~Studies} \label{sec:case_studies}
In this section, seven case studies have been conducted to depict the response of the proposed~CAE-HIFD.

\subsubsection{Case Study I---Close-in~HIF}

Figure~\ref{fig:HIFclosein} illustrates the performance of the CAE-HIFD under both normal and HIF conditions. In~this case study, the~HIF is applied at Node 632 starting at 0.05 s, as~seen in Figure~\ref{fig:HIFclosein}a. The~input voltage and current signals observed at the substation relay are shown in Figure~\ref{fig:HIFclosein}b,c, and~the kurtosis calculated from those voltages and currents is displayed in Figure~\ref{fig:HIFclosein}d. During~normal operation, the~kurtosis is below the threshold; upon the HIF inception, it raises over the threshold for approximately 8--10 ms returning quickly back to below threshold values. The~HIF causes the CC value to rise above the threshold, Figure~\ref{fig:HIFclosein}e, and, therefore, a~trip signal is issued approximately 60 ms after the HIF inception as seen in Figure~\ref{fig:HIFclosein}f.

\begin{figure}[H]
%\centering
\includegraphics[width=10.5 cm]{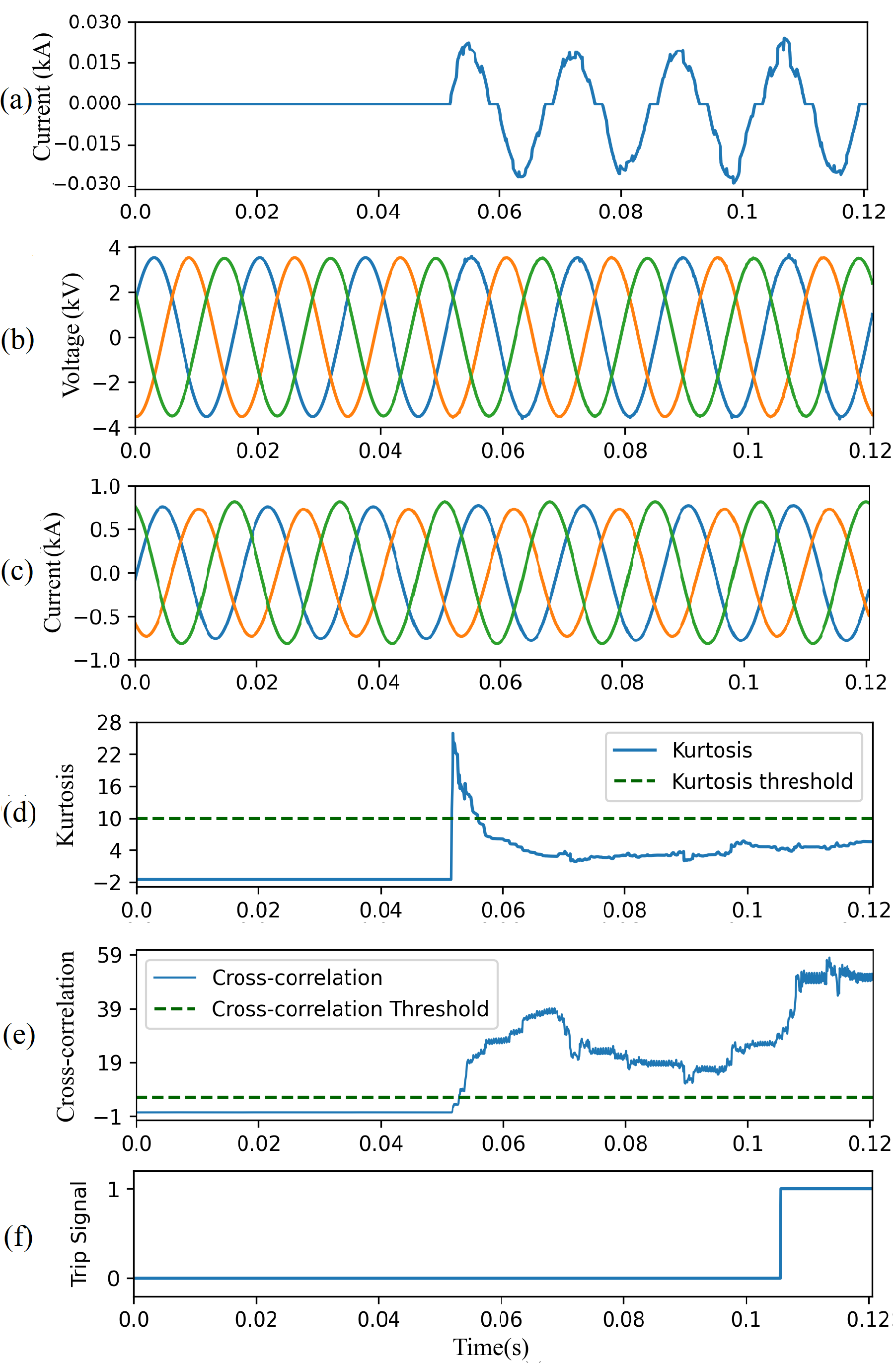}
\caption{CAE-HIFD performance under normal and close-in HIF conditions: (\textbf{a}) HIF current,\linebreak (\textbf{b}) three-phase input voltages, (\textbf{c}) three-phase input currents, (\textbf{d}) kurtosis, (\textbf{e}) cross-correlation, and~(\textbf{f}) trip~signal.\label{fig:HIFclosein}}
\end{figure}
\unskip

\subsubsection{Case Study II---Remote~HIF}
\textls[-25]{Figure~\ref{fig:HIFres} depicts the result of the CAE-HIFD in presence of a remote HIF: the HIF is applied at Node 652 starting at 0.05 s, as~seen is Figure~\ref{fig:HIFres}a. The~input voltage and current signals are observed in Figure~\ref{fig:HIFres}b,c, and~the calculated kurtosis is shown in Figure~\ref{fig:HIFres}d. Due to the remote location of HIF, the~HIF influence on the voltages and current signals is highly attenuated. As~a result, the~kurtosis surpasses the threshold for shorter duration of time (approximately 1--2 ms) remove as compared to Case study I. As~shown in Figure~\ref{fig:HIFres}e, the~CC value raises above the threshold after the inception of HIF. Consequently, a~trip signal is issued approximately 50 ms after the HIF inception as seen in Figure~\ref{fig:HIFres}f.}

\begin{figure}[H]
%\centering
\includegraphics[width=10.5 cm]{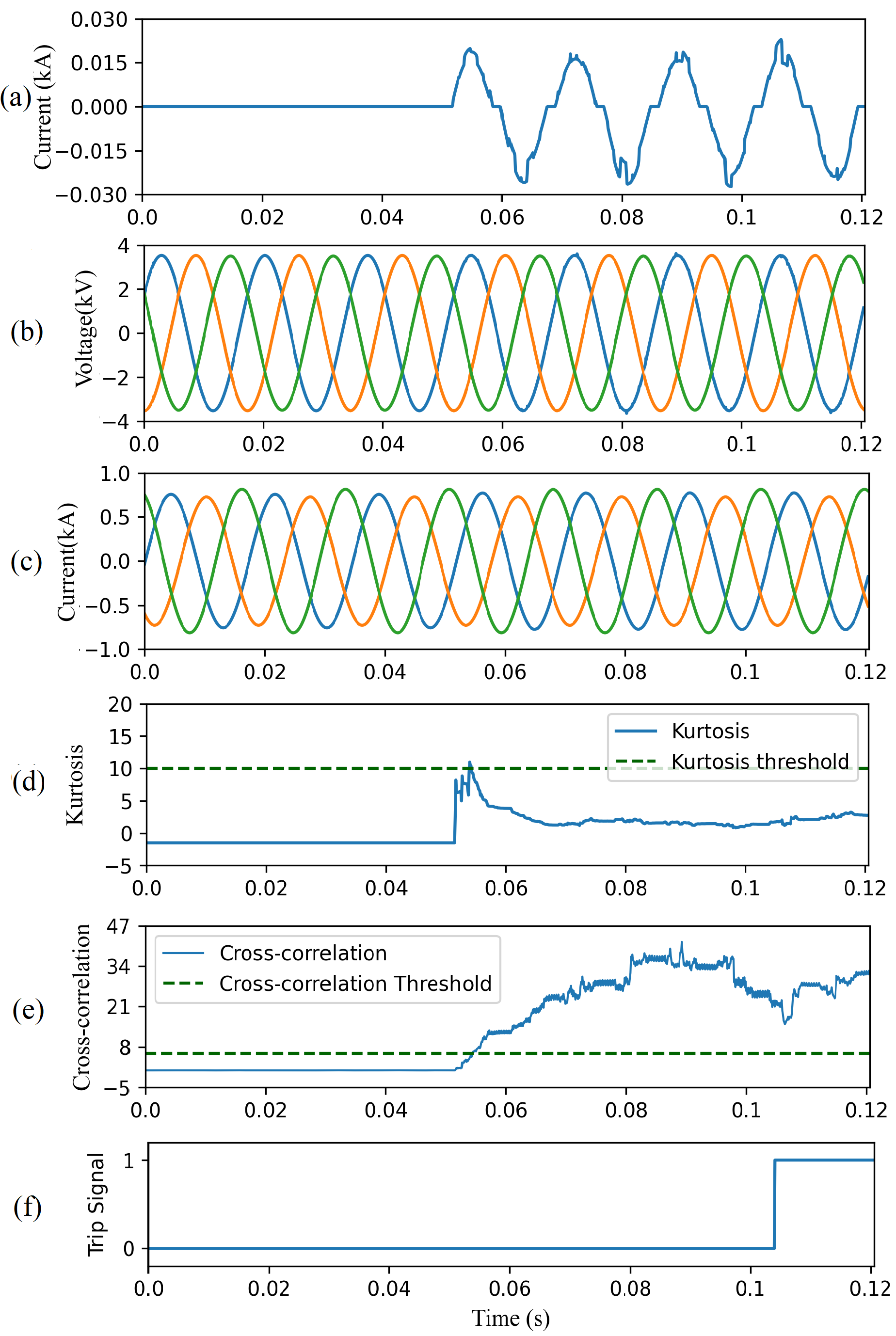}
\caption{CAE-HIFD performance under normal and remote HIF conditions: (\textbf{a}) HIF current,\linebreak (\textbf{b}) three-phase input voltages, (\textbf{c}) three-phase input currents, (\textbf{d}) kurtosis, (\textbf{e}) cross-correlation, and~(\textbf{f}) trip~signal.\label{fig:HIFres}}
\end{figure}
\unskip

\subsubsection{Case Study III---Capacitor~Switching}

The proposed HIF detection method successfully discriminates HIFs from switching events as demonstrated in Figure~\ref{fig:discap} with a three-phase capacitor bank located at node 675. Figure~\ref{fig:discap}a depicts phase A current caused by the capacitor energization at t = 0.05~s. The~current and voltage signals seen by the relay at the substation exhibit significant oscillations, as shown in Figure~\ref{fig:discap}b,c. This switching event causes sudden increase in the kurtosis for a short duration of time, approximately 15 ms (Figure~\ref{fig:discap}d). Although~the CC for the switching event is higher than its threshold, Figure~\ref{fig:discap}e, this disturbance is not falsely identified as an HIF, due to the high kurtosis value. Moreover, the~CC for the remaining non-HIF signal is below the threshold. Consequently, a~trip signal is not issued throughout the switching event, Figure~\ref{fig:discap}f.

\begin{figure}[H]
%\centering
\includegraphics[width=10.5 cm]{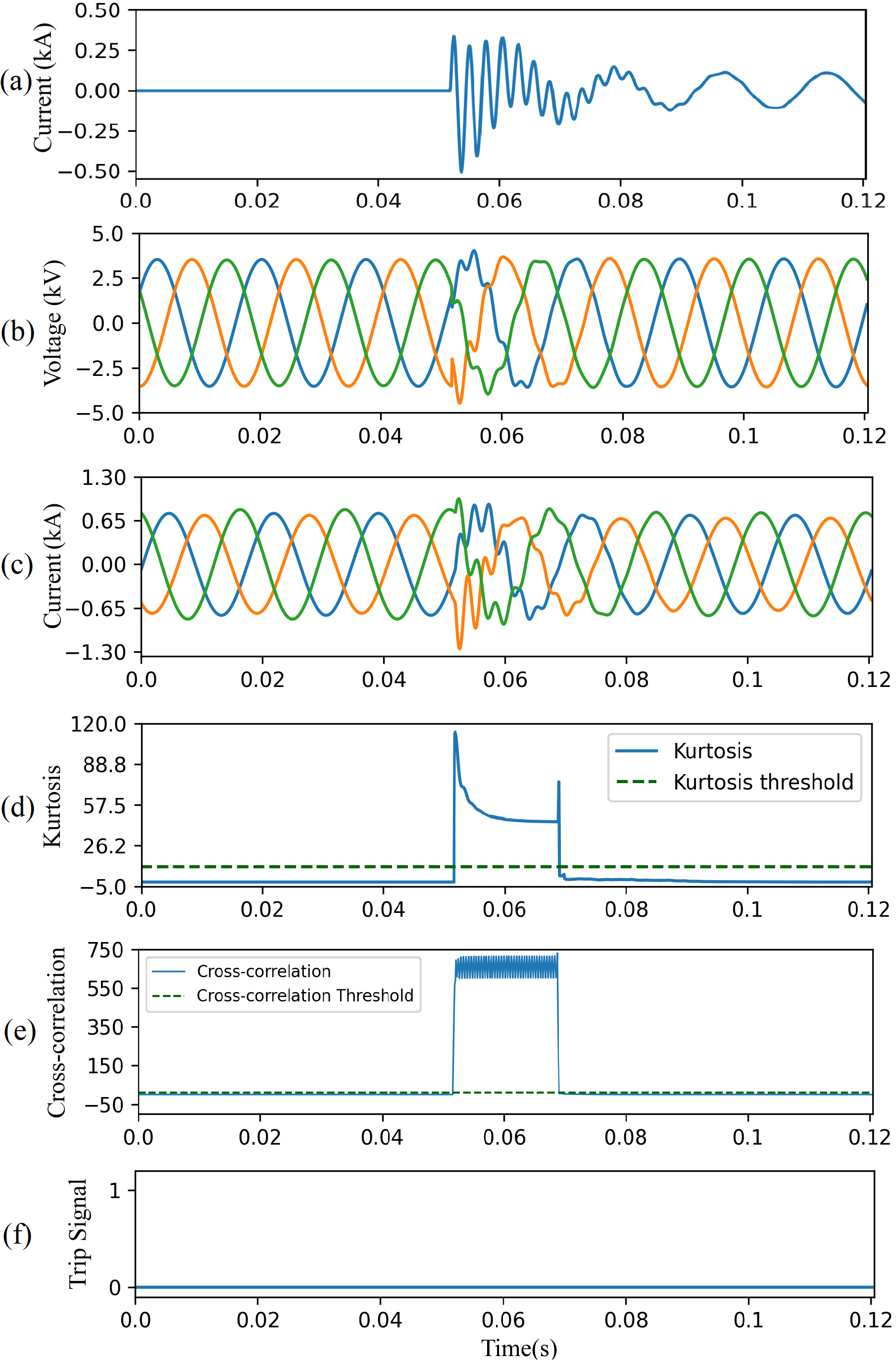}
\caption{CAE-HIFD performance under a capacitor switching scenario: (\textbf{a}) capacitor current,\linebreak (\textbf{b}) three-phase input voltages, (\textbf{c}) three-phase input currents, (\textbf{d}) kurtosis, (\textbf{e}) cross-correlation, and~(\textbf{f}) trip~signal.\label{fig:discap}}
\end{figure}
\unskip

\subsubsection{Case Study IV---Non-linear~Load}

Figure~\ref{fig:nonlinr} shows the performance of the proposed CAE-HIFD in presence of a non-linear load which causes significant harmonics. The~load at node 634 is replaced by a DC motor fed by a six-pulse thyristor rectifier. The~motor is started at t = 0.05 s. Figure~\ref{fig:nonlinr}a illustrates the phase A current of the non-linear load, while Figure~\ref{fig:nonlinr}b,c show voltages and currents measured by the relay at the substation. Although~the CC is higher than its threshold (Figure \ref{fig:nonlinr}e), the~trip signal (Figure \ref{fig:nonlinr}f) is not issued because the kurtosis surpasses its threshold (Figure \ref{fig:nonlinr}d).

\begin{figure}[H]
%\centering
\includegraphics[width=10.5 cm]{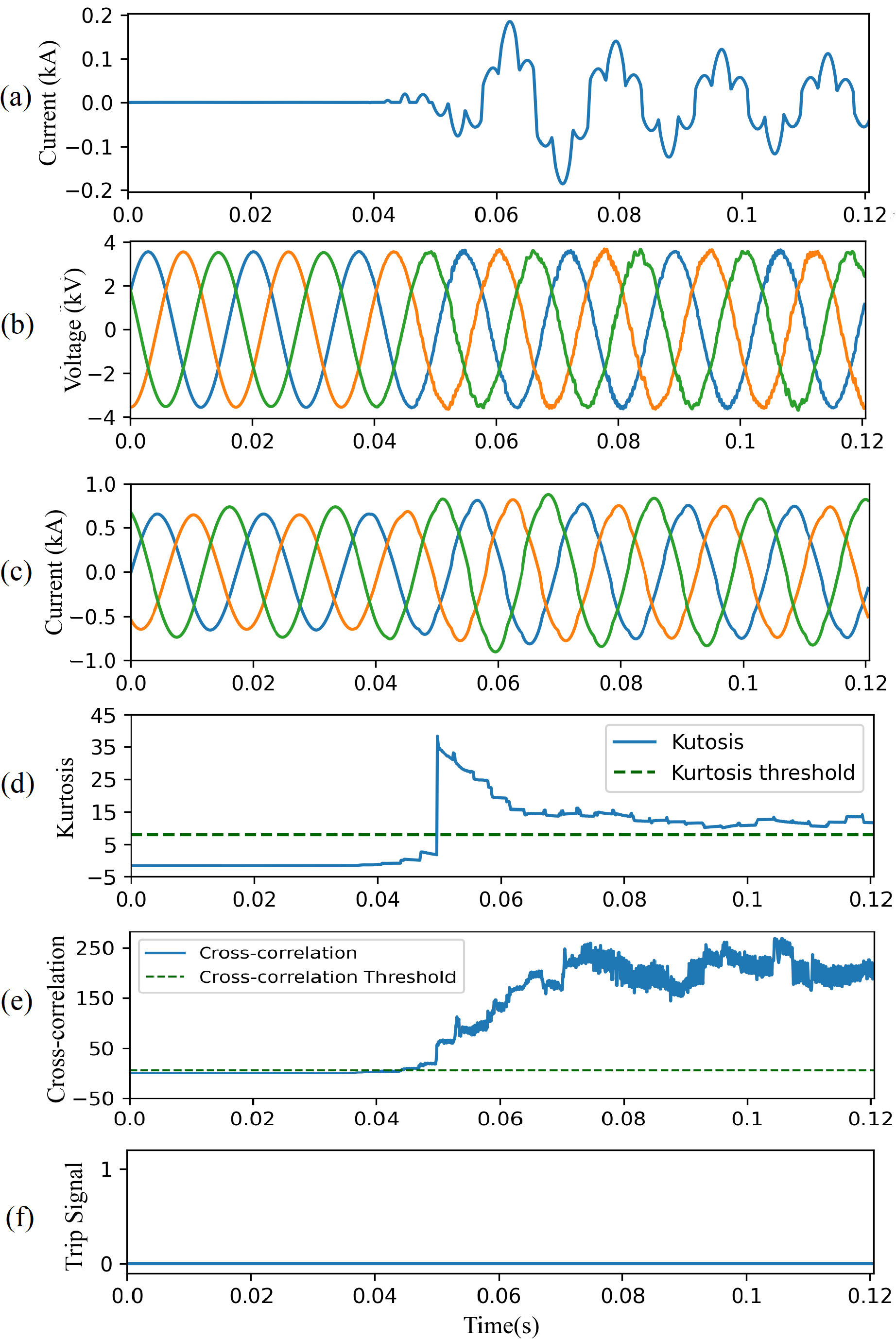}
\caption{CAE-HIFD performance under non-linear load switching: (\textbf{a}) non-linear load current,\linebreak (\textbf{b}) three-phase input voltages, (\textbf{c}) three-phase input currents, (\textbf{d}) kurtosis, (\textbf{e}) cross-correlation, and~(\textbf{f}) trip~signal.\label{fig:nonlinr}}
\end{figure}
\unskip

\subsubsection{Case Study V---Transformer~Energization}

This case study investigates the performance of the CAE-HIFD under a transformer energization scenario: the transformer at node 633 is energized at t = 0.05 s. The~inrush current for phase-A is shown in Figure~\ref{fig:inrush}a while Figure~\ref{fig:inrush}b,c display voltages and currents measured at the substation. Both the resulting kurtosis shown in Figure~\ref{fig:inrush}d, and~the CC shown in Figure~\ref{fig:inrush}e are below their corresponding thresholds. As~a result, the~proposed protection strategy does not cause any unnecessary tripping (Figure \ref{fig:inrush}f) under the transformer energization~scenario.

\begin{figure}[H]
%\centering
\includegraphics[width=10.5 cm]{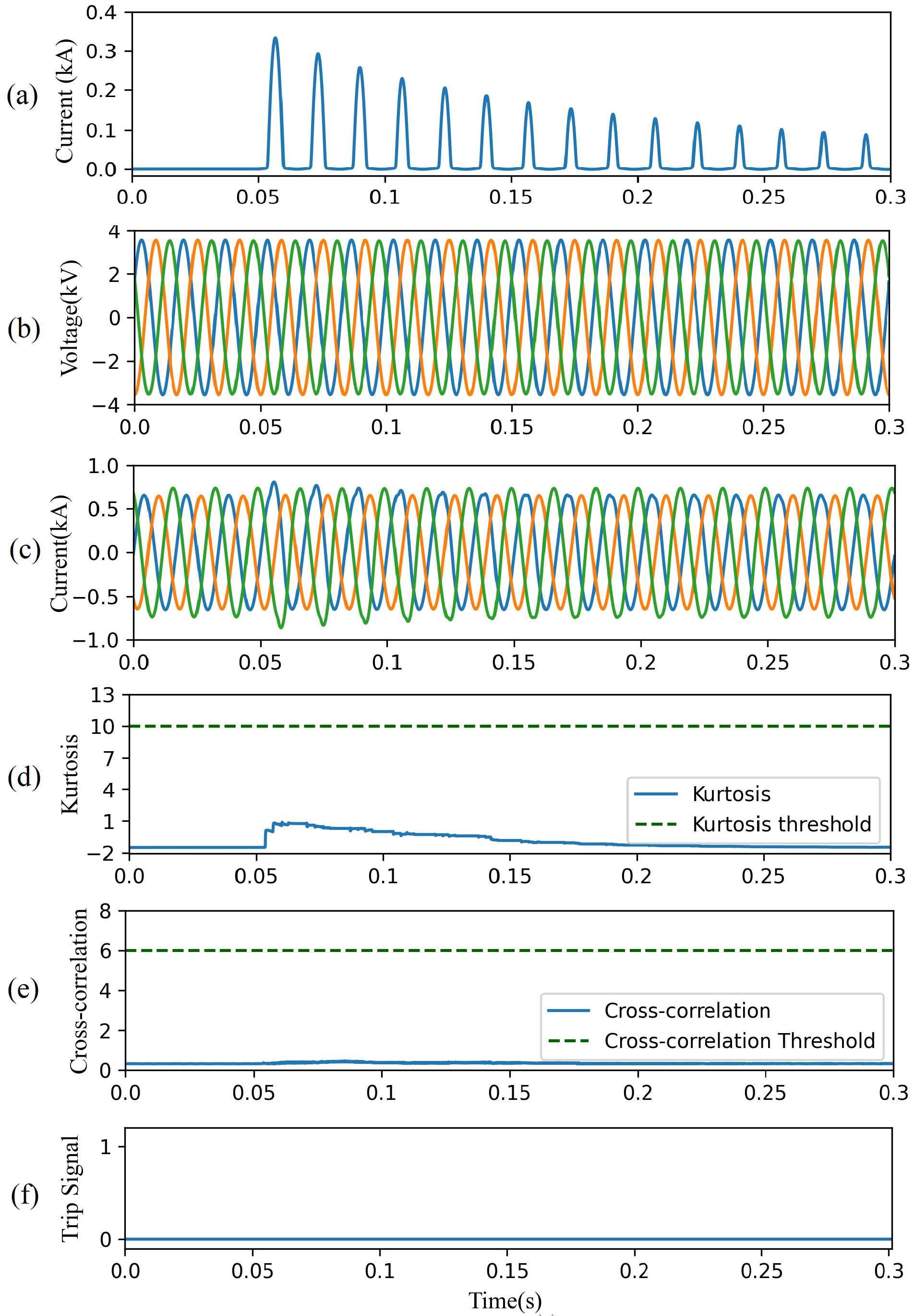}
\caption{CAE-HIFD performance under inrush currents: (\textbf{a}) Inrush current, (\textbf{b}) three-phase input voltages, (\textbf{c}) three-phase input currents, (\textbf{d}) kurtosis, (\textbf{e}) cross-correlation, and~(\textbf{f}) trip~signal.\label{fig:inrush}}
\end{figure}
\unskip

\subsubsection{Case Study VI---Intermittent~HIFs}

This case study demonstrates the effectiveness of the proposed CAE-HIFD in detecting intermittent HIFs. A~tree branch momentarily connects the phase-A to the ground for approximately 3.5 cycles (55 ms) as illustrated in Figure~\ref{fig:imfault}a. The~voltage and current signals shown in Figure~\ref{fig:imfault}b,c are measured by the relay at the substation. As~depicted in Figure~\ref{fig:imfault}d, the~kurtosis does not exceed the threshold. The~CC in Figure~\ref{fig:imfault}e crosses the  threshold during the intermittent faults. As~shown in Figure~\ref{fig:imfault}f, the~trip signal is issued after 50 ms. The~trip signal is reset after the intermittent fault is~cleared.

\begin{figure}[H]
%\centering
\includegraphics[width=10.5 cm]{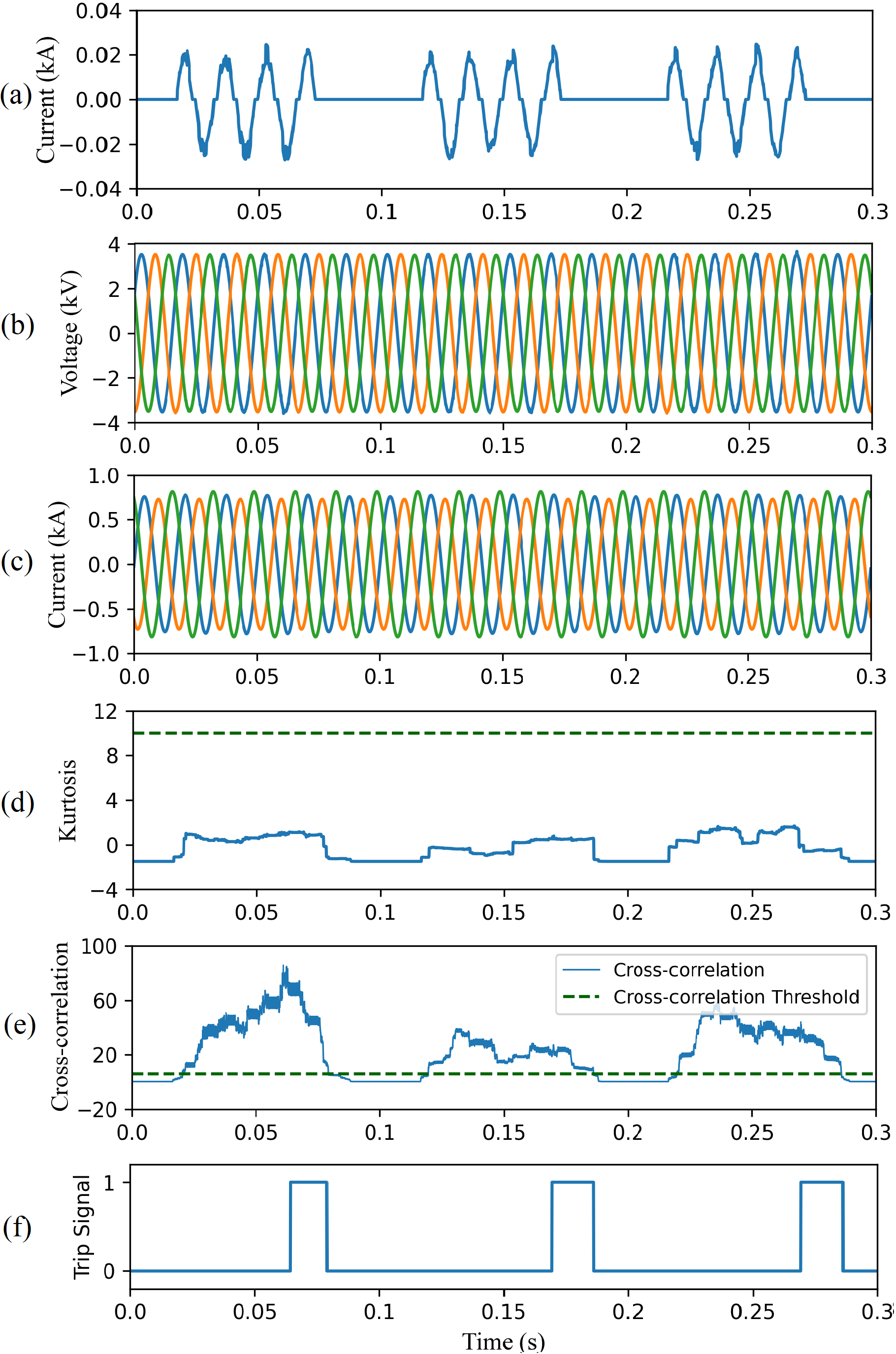}
\caption{CAE-HIFD performance under intermittent HIF condition: (\textbf{a}) fault current, (\textbf{b}) three-phase input voltages, (\textbf{c}) three-phase input currents, (\textbf{d}) kurtosis of the input, (\textbf{e}) cross-correlation, and~(\textbf{f}) trip~signal.\label{fig:imfault}}
\end{figure}
\unskip

\subsubsection{Case Study VII---Frequency~Deviations} \label{subsec:freq}

To demonstrate the effectiveness of the proposed method in presence of  frequency deviations, the~system frequency is increased to 61 Hz in this case study. The~HIF is initiated at t = 0.05 s. Figure~\ref{fig:freqdev}a,b represent currents and voltages measured by the relay at the substation. As~shown in Figure~\ref{fig:freqdev}c, before~the HIF takes place, the~kurtosis is below the threshold. As~the the HIF samples enters the sliding window, the~kurtosis exceeds the threshold because the distribution suddenly changes during the transition. Next, the~kurtosis returns to values below the threshold. The~CC in Figure~\ref{fig:freqdev}d is above the CC threshold; therefore, the~system trips within three cycles of the HIF~inception.

\begin{figure}[H]
%\centering
\includegraphics[width=10.5 cm]{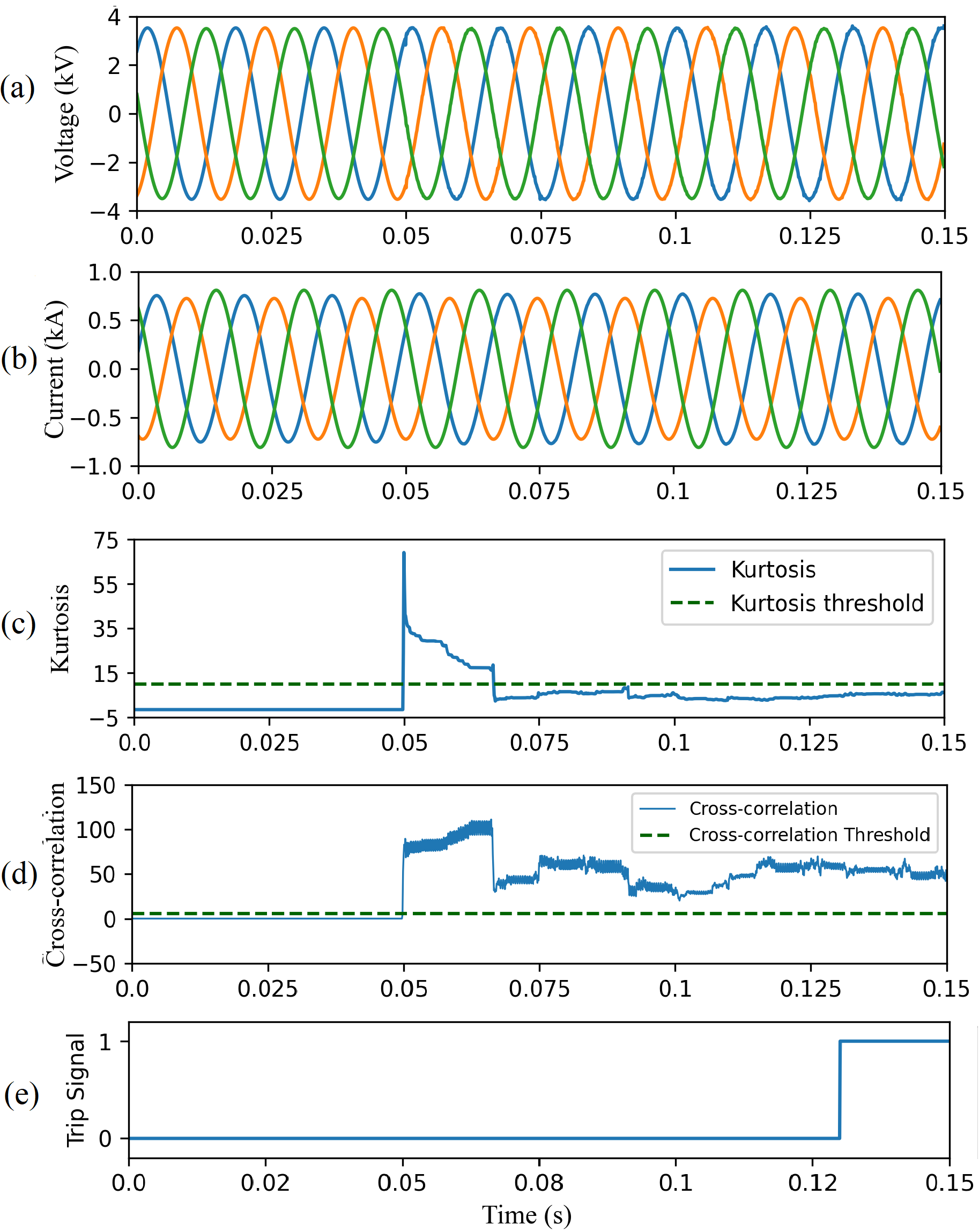}
\caption{CAE-HIFD performance under 61 Hz fundamental frequency: (\textbf{a}) three-phase input voltages, (\textbf{b}) three-phase input currents, (\textbf{c}) kurtosis, (\textbf{d}) cross-correlation, and~(\textbf{e}) trip~signal.\label{fig:freqdev}}
\end{figure}
\unskip

% \vspace{-8pt}
\subsection{Comparison with Other~Approaches} \label{sec:results}
\textls[-25]{This section first compares the proposed CAE-HIFD with other supervised and unsupervised learning algorithms. The~two supervised models selected for the comparison are: support vector machine (SVM) \cite{pow4} and artificial neural network (ANN) \cite{ANN}. As~supervised models require the presence of both, HIF and non-HIF data in the training set, these models are trained with a dataset containing an equal number of the HIF and non-HIF instances. Moreover, as~those models have originally been used with the DWT applied on the current waveform~\cite{pow4, ANN}, DWT is used here too. DWT extracts features by decomposing each phase current into seven detail level coefficients and one approximate level coefficient using the db4 mother wavelet. The~features are \linebreak formed by computing the standard deviation of coefficients at each level; therefore, eight standard deviations from each phase form a new input sample with 24 elements~\cite{ANN}. As~with CAE-HIFD, the~SVM and ANN hyperparameters are tuned using GDCV. The~SVM kernel is RBF with $\gamma$  of 0.05. The~ANN has three layers with 24-18-1 neurons, the~activation function for input and hidden layers is ReLU, and~binary cross-entropy is the loss~function.}

As other studies have only used supervised learning, to~examine unsupervised learning techniques, variations of the proposed approach are considered in this evaluation. Figure~{\ref{simpleFramework}} shows the flowchart for the unsupervised ML models. Preprocessing, kurtosis, and~CC calculation components are exactly the same as in the proposed CAE-HIFD, while two options are considered for the autoencoder algorithm and the training dataset. As~the autoencoder, the~proposed CAE-HIFD uses CAE while here we consider a variant of recurrent neural network, gated recurrent units autoencoder (GRU-AE). GRU-AE is selected because it is successful in extracting patterns from time-series data such as those present in the current and voltage signals. The~GRU-AE tuned with GDCV has two hidden layers, each one with 32 GRU cells and the ReLU activation function. For~both, CAE-HIFD and GRU-AE, two types of studies are conducted: training on HIFs data only and training on non-HIFs data~only.

\begin{figure}[H]
%\centering
\includegraphics[width=10.5 cm]{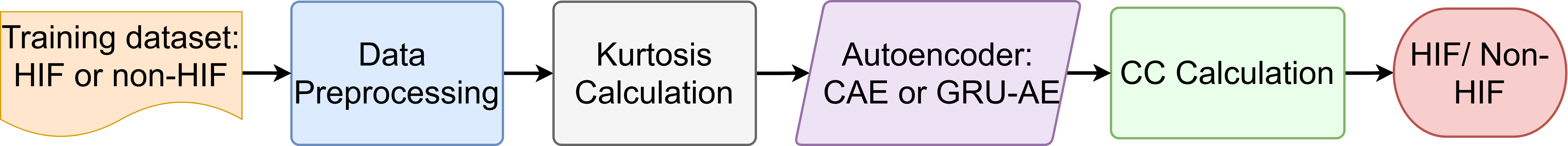}
\caption{Flowchart for the alternative unsupervised HIF detection models.
\label{simpleFramework}}
\end{figure}

The results of the comparison between CAE-HIFD and the other approaches are shown in Table~\ref{tab1}. It can be observed that CAE-HIFD outperforms other approaches and is the only one not susceptible to false tripping as indicated by the security metrics. In~addition, the~CAE-HIFD trained only with HIF data is highly efficient in discriminating non-HIF instances by detecting deviations from the learned HIF patterns. This prevents the algorithm from false tripping in the case of a new non-HIF pattern not present in the training set. The~results of the studies presented in Table~\ref{tab1} indicate that the CAE-HIFD achieves equally good results regardless of whether it is trained on the HIF data or non-HIF data; however, when trained with non-HIF data, there is a risk of identifying new non-HIF patterns as HIFs. The~supervised learning-based approaches category can only recognize the pattern present in the training set and, thus, may recognize new non-HIF events as HIFs. Overall, the~proposed CAE-HIFD achieves better performance than the other approaches.
% \vspace{-5pt}

% The MDPI table float is called specialtable
\begin{specialtable}[H]
\setlength{\tabcolsep}{2.9mm}
%\centering
\caption{Comparison of CAE-HIFD with other HIF detection~approaches.\label{tab1}}
%%% \tablesize{} %% You can specify the fontsize here, e.g.,~\tablesize{\footnotesize}. If commented out \small will be used.
\begin{tabular}{lccccc}
\toprule
\textbf{Model} & \textbf{Acc} & \textbf{Saf} & \textbf{Sen} & \textbf{Sec} & \textbf{Dep} \\

\midrule
Other Approaches-Supervised %is bold necessary? If not please remove it from table body.
& & & & &  \\
\hspace*{1em} DWT+SVM~\cite{pow4} & 97.97 & 100 &  97.78 & 78.99 & 100 \\
\hspace*{1em} DWT+ANN~\cite{ANN} & 97.72 & 100 & 97.55 & 76.47 & 100 \\

\midrule
Variants of our approach-Unsupervised & & & & &  \\
\hspace*{1em} Train on non-faults & & & & &  \\
\hspace*{3em} GRU-AE & 99.92 & 100 &  96.92 & 99.92 & 100 \\
\hspace*{3em} Proposed CAE-HIFD & 100 & 100 & 100 & 100 & 100 \\
\hspace*{1em} Train on faults & & & & &  \\
\hspace*{3em} GRU-AE & 34.61 & 32.24 & 83.22 & 97.52 & 5.66 \\
\hspace*{3em} Proposed CAE-HIFD & 100 & 100 & 100 & 100 & 100 \\

\bottomrule
\end{tabular}
\end{specialtable}

\subsection{Robustness of the Proposed CAE-HIF against~Noise} \label{sec:noise}

To examine CAE-HIFD robustness against noise, studies are conducted by introducing different levels of noise. The~white Gaussian noise is considered because it covers a large frequency spectrum. The~noise is added to the current signals because current waveforms are more susceptible to noise~\cite{pow1}. As~shown in Figure~\ref{fig:noise}, the~proposed CAE-HIFD approach is immune to noise when signal to noise ratio (SNR) value is higher than 40 dB. In~a case of high nose, SNR below 40 dB, the~accuracy reduces to 97\%. Thus, more than one consecutive window is needed to be processed before making a tripping decision in order to avoid undesired tripping and to ensure accurate HIF detection. Therefore, three consecutive windows are utilized in all performance evaluation studies in order to accurately detect all HIFs. Increasing the timer threshold improves the resiliency against unnecessary tripping, but~prolongs HIF detection time. Even with the extremely noisy condition of 1 dB SNR, the~accuracy, security, and~sensibility  do not fall below 97.65\%, 94.03\%, 96.28\%, respectively. The~reason behind this robustness is the CAE de-noising ability and strong pattern learning capability. The~inherent de-noising nature of the autoencoders assists the CAE to generalize the corrupted input. Additionally, the~CAE-HIFD learns the complex HIF patterns because of spatial feature learning proficiency of the CAE. The~model accurately detects non-HIF scenarios under considered noise levels; hence, the~safety and dependability remains at 100\% even with high levels of noise. Moreover, the~values of other performance metrics are also greater than 90\% throughout the SNR range of 5dB to 50dB demonstrating noise robustness of the~CAE-HIFD.

Figure~\ref{fig:noiseResp} shows CAE-HIFD performance in presence of noise of 20 dB SNR. Before~the HIF inception at 0.05 s, despite the significant noise, the~kurtosis (Figure \ref{fig:noiseResp}b) and the CC (Figure \ref{fig:noiseResp}c) remain below their thresholds. The~CC surpasses the threshold upon HIF inception and, as~a result, the~designed protection system issues trip signal (Figure \ref{fig:noiseResp}d).

\begin{figure}[H]
\includegraphics[width=10.5 cm]{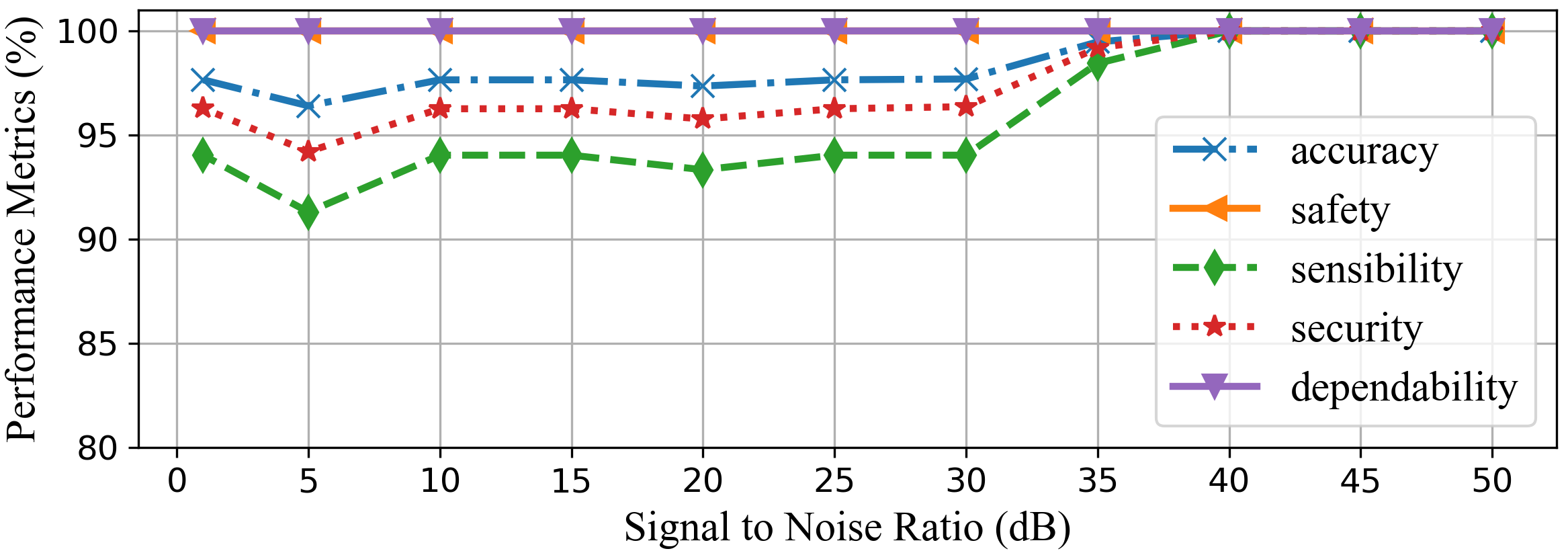}
%\centering
\caption{Effect of noise on the CAE-HIFD~performance.\label{fig:noise}}
\end{figure}
\unskip

\begin{figure}[H]
%\centering
\includegraphics[width=10.5 cm]{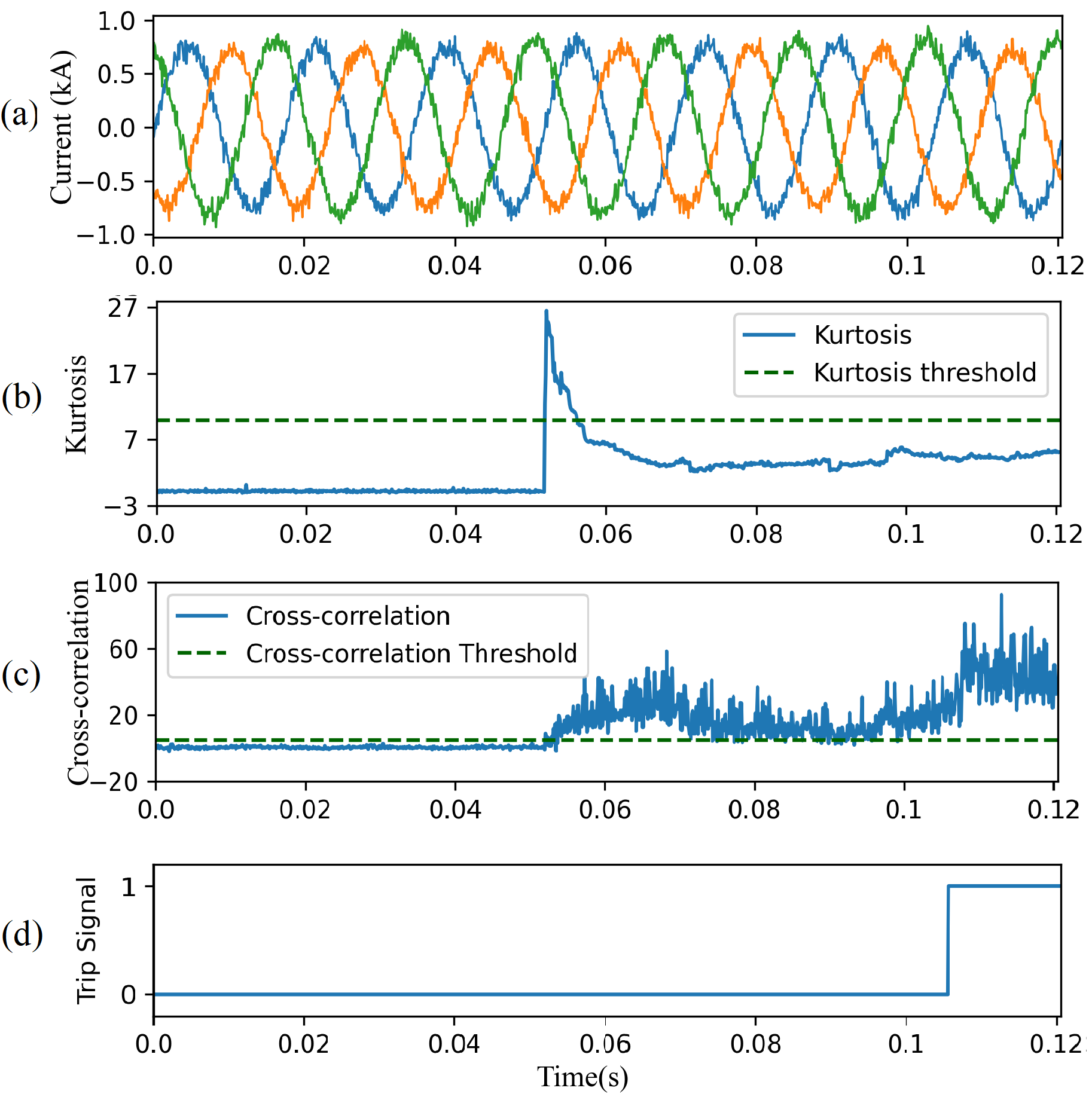}
\caption{CAE-HIFD performance under 20 dB SNR: (\textbf{a}) three-phase input currents, (\textbf{b}) kurtosis,\linebreak (\textbf{c}) cross-correlation, and~(\textbf{d}) trip~signal.\label{fig:noiseResp}}
\end{figure}
\unskip

\subsection{Discussion}

{The evaluation results demonstrate that the proposed CAE-HIFD achieves 100\% HIF detection accuracy irrespective of the surface type, fault phase, and~fault location. All~metrics, including accuracy, safety, sensibility, security, and~dependability are at 100\% as shown in Table {\ref{tab2}}. Moreover, for~all considered scenarios, the~system trips within three cycles after the HIF~inception.

The challenging part of machine learning for HIF detection is in the diversity of non-fault and fault signals together with similarities between non-HIF disturbances and HIFs. By~training on faults only, the~proposed approach does not require simulation of non-fault scenarios for training. Distinguishing HIFs from the non-HIF steady-state operation can take advantage of the smoothness of non-HIF steady-state signal; however, non-HIF disturbances, such as capacitor and load switching, share many characteristics (e.g., randomness and non-linearity) with the HIF signals making it difficult for a neural network (in our case CAE) to distinguish between them. To~address this challenge, the~proposed approach takes advantage of differences in data distributions between non-fault disturbances and HIFs and employs kurtosis to differentiate between the~two.

In experiments, 210 fault cases were considered corresponding to 1372 fault data windows as described in Section {\ref{sec:studies}}. Signals corresponding to these faults are different from each other as simulations included different surfaces, fault locations, and~fault phases. From~these fault cases, 80\% is selected randomly for training, therefore, some cases are present only in testing. Moreover, all the case studies presented in Section {\ref{sec:case_studies}} are conducted with data that are not seen by the proposed CAE-HIFD in training. The~proposed system successfully distinguished between fault and non-fault signals for all scenarios, which demonstrates its abilities to detect previously unseen HIF and non-HIF~scenarios.

Frequency deviations, as~well as noise, impose major challenges for the HIF detection. Approaches that operate on the fundamental frequency components risk failures in presence of frequency deviations. However, CAE-HIFD does not operate based on the fundamental frequency components of the input signals and, consequently, is not sensitive to frequency deviations as shown in Section {\ref{subsec:freq}}. As~noise is common in distribution systems, it is important to consider it in HIF detection evaluation. HIF detection in presence of noise is difficult as noisy signals are accompanied by randomness and have characteristics that resemble HIFs. Nonetheless, experiments from Section {\ref{sec:noise}} show that CAE-HIFD remains highly accurate even in presence of significant noise.}

\section{Conclusions} \label {sec:conclusion}

Recently, various machine learning-based methods have been proposed to detect HIFs. However, these methods utilize supervised learning; thus, they are prone to misclassification of HIF or non-HIF scenarios that are not present in the training~data.

This paper proposes the CAE-HIFD, a~novel deep learning-based approach for HIF detection capable of reliably discriminating HIFs from non-HIF behavior including diverse disturbances. The~convolutional autoencoder in CAE-HIFD learns from the fault data only, which eliminates the need of considering all possible non-HIF scenarios for the training process. The~MAE commonly used to compare autoencoder input and output is replaced by cross-correlation in order to discriminate HIFs from disturbances such as capacitor and load switching. To~distinguish switching events from HIFs, the~CAE-HIFD employs kurtosis~analysis.

The results show that CAE-HIFD achieves 100\% performance in terms of all five metrics of protection system performance, namely accuracy, security, dependability, safety, and~sensitivity. The~proposed CAE-HIFD outperforms supervised learning approaches, such as the SVM with DWT and the ANN with DWT, as~well as the unsupervised GRU-based autoencoder. The~CAE-HIFD performance is demonstrated on case studies including steady-state operation, close-in and remote HIFs, capacitor switching, non-linear load, transformer energization, intermittent faults, and~frequency deviations. The~studies on the effect of different noise levels demonstrate that the proposed CAE-HIFD is robust against noise for SNR levels as low as 40 dB and provides acceptable performance for higher noise~levels.

Future work will examine HIF detection from only voltage or on current signals in order to reduce computational complexity. Furthermore, HIF classification technique will be developed to determine the phase on which the fault~occurred.

% \section*{Acknowledgements}

% This research has been supported by NSERC under grants RGPIN-2018-06222 and RGPIN-2017-04772.

\vspace{6pt}

\authorcontributions{Conceptualization, K.R.; methodology, K.R.; software, K.R. and F.H.; validation, K.R. and F.H.; formal analysis, K.R., F.H., K.G. and F.B.A.; investigation, K.R. and F.H.; resources, K.G. and F.B.A.; data curation, F.H.; writing---original draft preparation, K.R. and F.H.; writing---review and editing, K.G., F.B.A., K.R., F.H.; visualization, K.R.; supervision, K.G. and F.B.A.; project administration, K.G. and F.B.A.; funding acquisition, K.G. and F.B.A. All authors have read and agreed to the published version of the~manuscript.}

\funding{This research has been supported by NSERC under grants RGPIN-2018-06222 and~RGPIN-2017-04772.}

%\institutionalreview{\hl{ }}%In this section, please add the Institutional Review Board Statement and approval number for studies involving humans or animals. Please note that the Editorial Office might ask you for further information. Please add ``The study was conducted according to the guidelines of the Declaration of Helsinki, and approved by the Institutional Review Board (or Ethics Committee) of NAME OF INSTITUTE (protocol code XXX and date of approval).'' OR ``Ethical review and approval were waived for this study, due to REASON (please provide a detailed justification).'' OR ``Not applicable'' for studies not involving humans or animals. You might also choose to exclude this statement if the study did not involve humans or animals.

%\informedconsent{\hl{ }}%Any research article describing a study involving humans should contain this statement. Please add ``Informed consent was obtained from all subjects involved in the study.'' OR ``Patient consent was waived due to REASON (please provide a detailed justification).'' OR ``Not applicable'' for studies not involving humans. You might also choose to exclude this statement if the study did not involve humans. Written informed consent for publication must be obtained from participating patients who can be identified (including by the patients themselves). Please state ``Written informed consent has been obtained from the patient(s) to publish this paper'' if applicable.

\dataavailability{The data are not publicly available due to privacy.}

%In this section, please provide details regarding where data supporting reported results can be found, including links to publicly archived datasets analyzed or generated during the study. Please refer to suggested Data Availability Statements in section ``MDPI Research Data Policies'' at \url{https://www.mdpi.com/ethics}. You might choose to exclude this statement if the study did not report any data.

\conflictsofinterest{The authors declare no conflict of~interest.}

%% Optional
% \sampleavailability{Samples of the compounds ... are available from the authors.}

%%%%%%%%%%%%%%%%%%%%%%%%%%%%%%%%%%%%%%%%%%
%% Only for journal Encyclopedia
%\entrylink{The Link to this entry published on the encyclopedia platform.}

%%%%%%%%%%%%%%%%%%%%%%%%%%%%%%%%%%%%%%%%%%
%% Optional
\abbreviations{Abbreviations}{
The following abbreviations are used in this manuscript:\\

\noindent
\begin{tabular}{@{}ll}
CAE-HIFD & Convolutional Autoencoder framework for HIF Detection\\
CAE &  Convolutional Autoencoder\\
HIF & High-Impedance Fault\\
EMD & Empirical Mode Decomposition\\
VMD &  Variational Mode Decomposition\\
ML &  Machine Learning\\
SVM & Support Vector Machine\\
SNR & Signal to Noise Ratio\\
DWT & Discrete Wavelet Transform\\
WT & Wavelet Transform\\
CNN & Convolutional Neural Network\\
CC & Cross-Correlation\\
MSE & Mean Squared Error\\
ReLU & Rectified Linear Unit\\
K & Kurtosis\\
Acc & Accuracy\\
Dep & Dependability\\
Saf & Safety\\
Sen & sensibility\\
TP & True Positives\\
TN & True Negatives\\
FN & False Negatives\\
FP & False Positives\\
GSCV & Grid Search Cross-Validation\\
ANN & Artificial Neural Network\\
GRU-AE & Gated Recurrent Units Autoencoder

\end{tabular}}

%%%%%%%%%%%%%%%%%%%%%%%%%%%%%%%%%%%%%%%%%%
%% Optional
\appendixtitles{no} % Leave argument "no" if all appendix headings stay EMPTY (then no dot is printed after "Appendix A"). If~the appendix sections contain a heading then change the argument to "yes".
\appendixstart

\appendix
\section{}\label{app:a}
% \subsection{}

% \hl{The utilized study system,} %Please delete Appendix A.1 it it's not necessary.
%  the~IEEE 13 node test feeder, is carrying significant levels of imbalance in a steady state. This system is chosen because it contains (i) three phases, two phases and single-phase overhead lines and underground cables, and (ii) is heavily loaded and unbalanced. Furthermore, the~load data in Table~\ref{tab4} shows the unbalanced nature of the utilized test~system.
\vspace{-10pt}

\begin{specialtable}[H]
\setlength{\tabcolsep}{2.85mm}
%\tablesize{\scriptsize}
%\centering
\caption{IEEE 13 node test feeder Load~Information.\label{tab4}}
%\tablesize{} % You can specify the fontsize here, e.g.,~\tablesize{\footnotesize}. If commented out \small will be used.
\tablesize{\footnotesize}

\begin{tabular}{ p{1cm}<\centering p{1cm}<\centering p{1.2cm}<\centering p{1.2cm}<\centering p{1.2cm}<\centering p{1.2cm}<\centering p{1.2cm}<\centering p{1.2cm}<\centering }
\toprule
\textbf{Node} & \textbf{Load Model}  & \textbf{Phase 1 (kW)} & \textbf{Phase 1  (kVar)} & \textbf{Phase 2  (kW)} & \textbf{Phase 2 (kVar)} & \textbf{Phase 3 (kW)} & \textbf{Phase 3 (kVar)} \\

\midrule
634 & Y-PQ & 160 & 110 & 120 & 90  & 120 & 90 \\
645 & Y-PQ & 0   & 0   & 170 & 125 &  0  & 0 \\
646 & D-Z  & 0   & 0  & 230  & 132 &  0  & 0 \\
652 & Y-Z  & 128 & 86 & 0   &  0  &  0  &  0 \\
671 & D-PQ & 385 & 220 & 385 & 220 & 385 & 220 \\
675 & Y-PQ & 485 & 190 & 68 & 60 & 290 & 212 \\
692 & D-I & 0 & 0 & 0 & 0 & 170 & 151 \\
611 & Y-I & 0 & 0 & 0 & 0 & 170 & 80 \\
& Total & 1158 & 606 & 973 & 627 & 1135 & 735 \\
\bottomrule
\end{tabular}
\end{specialtable}

\begin{specialtable}[H]
\setlength{\tabcolsep}{9.62mm}
%\centering
%\tablesize{\scriptsize}
\caption{IEEE 13 node test feeder Line Length and~phasing.\label{tab5}}
%\tablesize{} % You can specify the fontsize here, e.g.,~\tablesize{\footnotesize}. If commented out \small will be used.

\begin{tabular}{cccccccc}
\toprule
\textbf{Node A}	& \textbf{Node B} & \textbf{Length (ft)} & \textbf{Phasing}\\

\midrule

632 & 645 & 500 &	C, B, N \\
632 &	633	& 500	& C, A, B, N \\
633	& 634 &	0 &	Transformer \\
645	& 646 & 300	& C, B, N \\
650	& 632 & 2000 & B, A, C, N \\
684 & 652 & 800	& A, N \\
632	& 671	& 2000	& B, A, C, N \\
671	& 684 & 300	& A, C, N \\
671	& 680	& 1000 & B, A, C, N \\
671	& 692 & 0 & Switch \\
684	& 611& 300 & C, N \\
692	& 675 & 500	& A, B, C, N \\

\bottomrule
\end{tabular}
\end{specialtable}

%%%%%%%%%%%%%%%%%%%%%%%%%%%%%%%%%%%%%%%%%%

%%%%%%%%%%%%%%%%%%%%%%%%%%%%%%%%%%%%%%%%%%
\end{paracol}
\reftitle{References}

\end{document}